\documentclass[10pt,journal,compsoc]{IEEEtran}

\usepackage{times}
\usepackage{epsfig}
\usepackage{graphicx}
\usepackage{amsmath}
\usepackage{amssymb}
\usepackage{wrapfig}
\usepackage[export]{adjustbox}
\usepackage{etoolbox}
\usepackage[normalem]{ulem}
\usepackage[symbol]{footmisc}
\usepackage[pagebackref=false,breaklinks=true,letterpaper=true,colorlinks,bookmarks=false]{hyperref}
\usepackage[normalem]{ulem}

\usepackage{stmaryrd}
\usepackage{placeins}

\usepackage{times}
\usepackage{epsfig}
\usepackage{graphicx}
\usepackage{float}
\usepackage{wrapfig}
\usepackage{amsmath,amssymb,amsthm}
\usepackage{algorithm,algorithmicx,algpseudocode}
\usepackage{bm,xspace}
\usepackage{comment}
\usepackage{verbatim}
\usepackage{multirow}
\usepackage{balance}
\usepackage{url}
\usepackage{booktabs}
\usepackage{etoolbox,siunitx}
\usepackage{calc}
\usepackage{pifont,hologo}
\usepackage[usenames, dvipsnames]{color}
\usepackage{nicefrac}

\newcommand{\ra}[1]{\renewcommand{\arraystretch}{#1}}

\usepackage[super]{nth}
\usepackage{nicefrac}
\robustify\bfseries

\def\dd{\mathbf{d}}

\def\hh{\mathbf{h}}

\def\zz{\mathbf{z}}

\def\lL{\mathcal{L}}

\def\Re{\mathbb{R}}

\def\btheta{{\bm\theta}}

\def\trans{^{\top}}

\newcommand{\YesV}{\ding{51}}%

\DeclareMathSymbol{@}{\mathord}{letters}{"3B}

\newcommand\abs[1]{\left\lvert#1\right\rvert}

\newcommand\mypara[1]{\vspace{1mm}\noindent\textbf{#1}}

\def\rot#1{\rotatebox{90}{#1}}

\def\latex/{\LaTeX}
\def\bibtex/{\hologo{BibTeX}}

\newcommand\changed[1]{{#1}}
\newcommand\removed[1]{}

\ifCLASSOPTIONcompsoc
  \usepackage[nocompress]{cite}
\else
  \usepackage{cite}
\fi

\begin{document}
\title{Towards Robust Monocular Depth Estimation:\\ Mixing Datasets for\\ Zero-shot Cross-dataset Transfer}

\author{Ren\'{e} Ranftl*, Katrin Lasinger*, David Hafner, Konrad Schindler, and Vladlen Koltun%
\IEEEcompsocitemizethanks{\IEEEcompsocthanksitem R. Ranftl, D. Hafner, and V. Koltun are with the Intelligent Systems Lab,
  Intel Labs.\protect\\
\IEEEcompsocthanksitem K. Lasinger and K. Schindler are with the Institute of Geodesy and Photogrammetry, ETH Z\"urich.}%
\thanks{*Equal contribution}}

\newcommand{\etal}{\MakeLowercase{\textit{et al.}}}
\newcommand{\ie}{\MakeLowercase{\textit{i.e.}}}
\newcommand{\eg}{\MakeLowercase{\textit{e.g.}}}
\newcommand{\cf}{\MakeLowercase{\textit{cf.}}}

\markboth{IEEE Transactions on Pattern Analysis and Machine Intelligence, ~Vol.~XX, No.~XX, ~2020}%
{Ranftl \MakeLowercase{\textit{et al.}}: Towards Robust Monocular Depth Estimation}

\IEEEtitleabstractindextext{%
\begin{abstract}
  The success of monocular depth estimation relies on large and diverse training sets.
  Due to the challenges associated with acquiring dense ground-truth depth across different environments at scale, a number
  of datasets with distinct characteristics and biases have emerged.
  We develop tools that enable mixing multiple datasets during training, even if their annotations are incompatible.
  In particular, we propose a robust training objective that is invariant to changes in depth range and scale, advocate the
  use of principled multi-objective learning to combine data from different sources, and highlight the importance of
  pretraining encoders on auxiliary tasks. Armed with these tools, we experiment with five diverse training datasets,
  including a new, massive data source: 3D films.
  To demonstrate the generalization power of our approach we use \emph{zero-shot cross-dataset transfer}, i.e.\ we evaluate
  on datasets that were not seen during training. The experiments confirm that mixing data from complementary sources
  greatly improves monocular depth estimation.
  Our approach clearly outperforms competing methods across diverse datasets, setting a new state of the art for monocular depth estimation.
\end{abstract}

}

\maketitle

\IEEEdisplaynontitleabstractindextext

\IEEEpeerreviewmaketitle

\IEEEraisesectionheading{\section{Introduction}\label{sec:introduction}}
\IEEEPARstart{D}{epth} is among the most useful intermediate representations for action in physical environments ~\cite{Zhou2019}. Despite its utility, monocular depth estimation remains a challenging problem that is heavily underconstrained. To solve it, one must exploit many, sometimes subtle, visual cues, as well as long-range context and prior knowledge. This calls for learning-based techniques~\cite{Hoiem2005,Saxena2009}.

\begin{figure*}[h!]
  \def\mycolspace{1.2mm}
  \def\teaserheight{3.47cm}
  \centering
 	\begin{tabular}{@{}c@{\hspace{\mycolspace}}c@{\hspace{\mycolspace}}c@{\hspace{\mycolspace}}c@{}}
    \includegraphics[height=\teaserheight,trim={0 0 0 0}, clip]{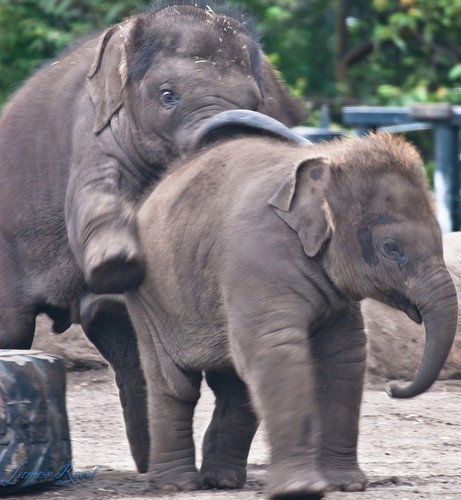} &
    \includegraphics[height=\teaserheight,trim={2.0cm 0 0 0}, clip]{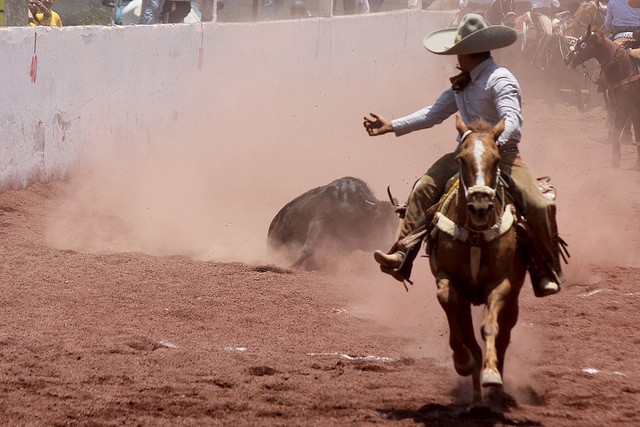} &
    \includegraphics[height=\teaserheight,trim={0 0 0 0}, clip]{figures/teaser_pcl/bags_in} &
    \includegraphics[height=\teaserheight,trim={0 0 0 0}, clip]{figures/teaser_pcl/bus_in} \\
      \includegraphics[height=\teaserheight,trim={0 0 0 0}, clip]{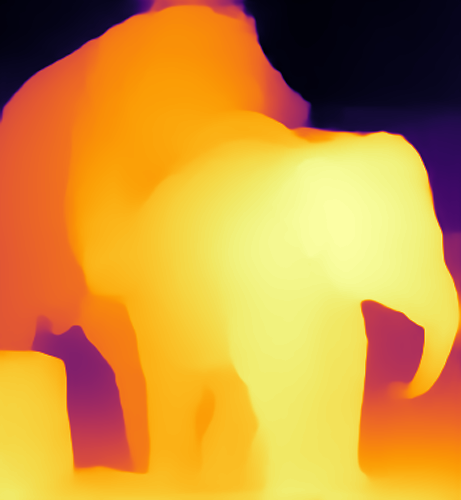} &
      \includegraphics[height=\teaserheight,trim={2.0cm 0 0 0}, clip]{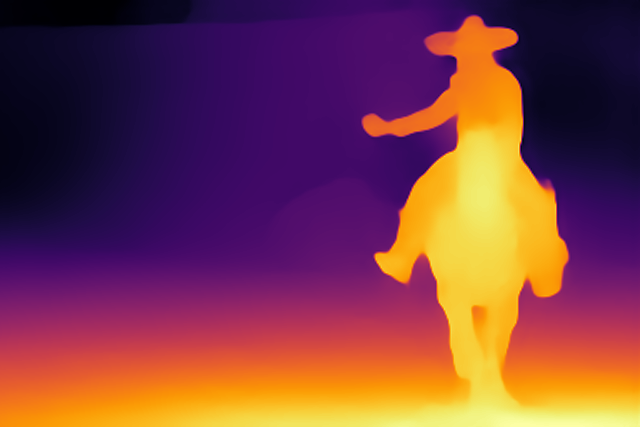} &
    \includegraphics[height=\teaserheight,trim={0 0 0 0}, clip]{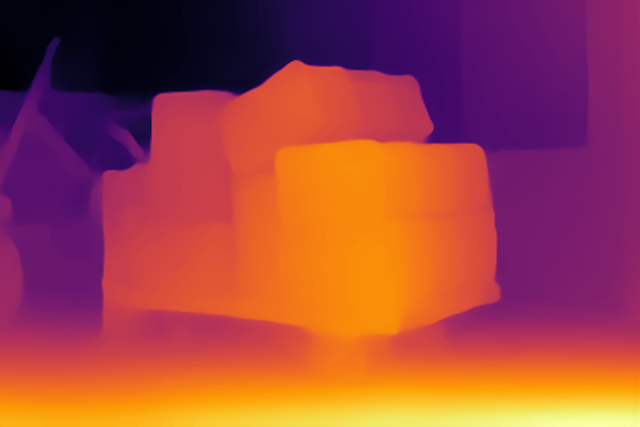} &
    \includegraphics[height=\teaserheight,trim={0 0 0 0}, clip]{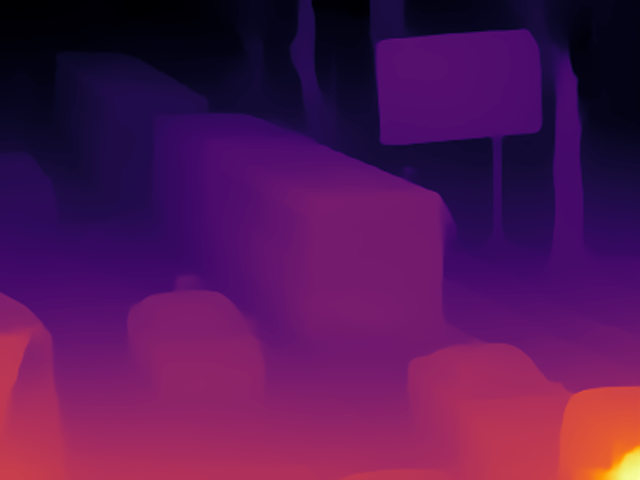} \\
      \includegraphics[height=\teaserheight,trim={2.75cm 0 1.75cm 0}, clip]{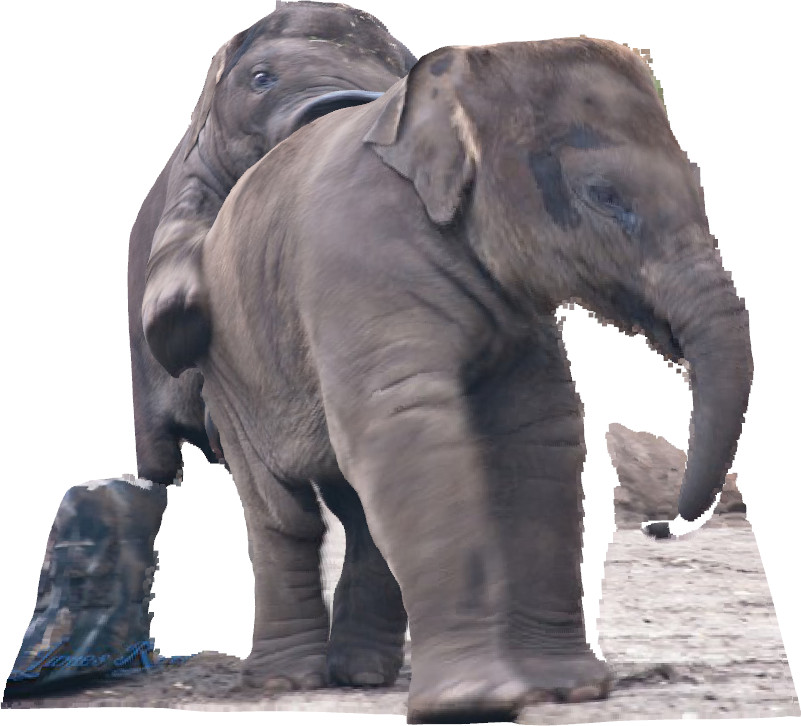} &
    \includegraphics[height=\teaserheight,trim={3cm 0 2cm 0}, clip]{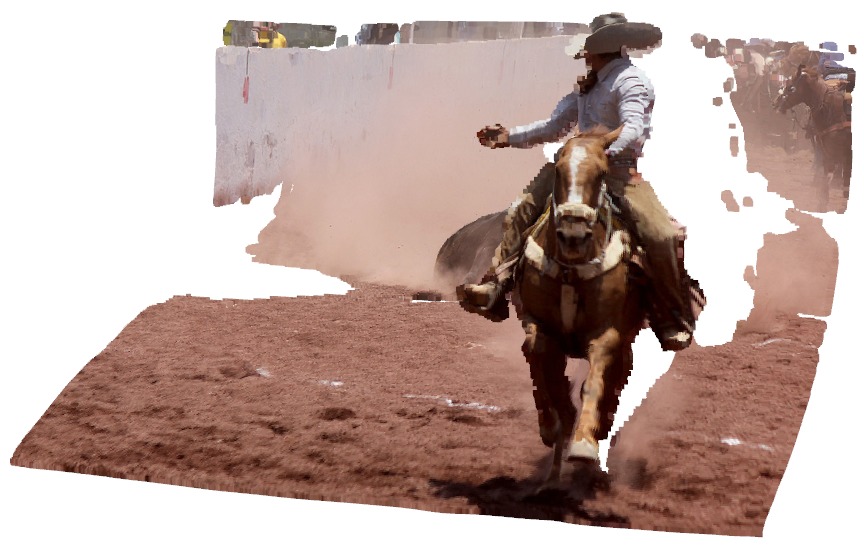} &
    \includegraphics[height=\teaserheight,trim={0cm 0.5cm 0.3cm 0cm}, clip]{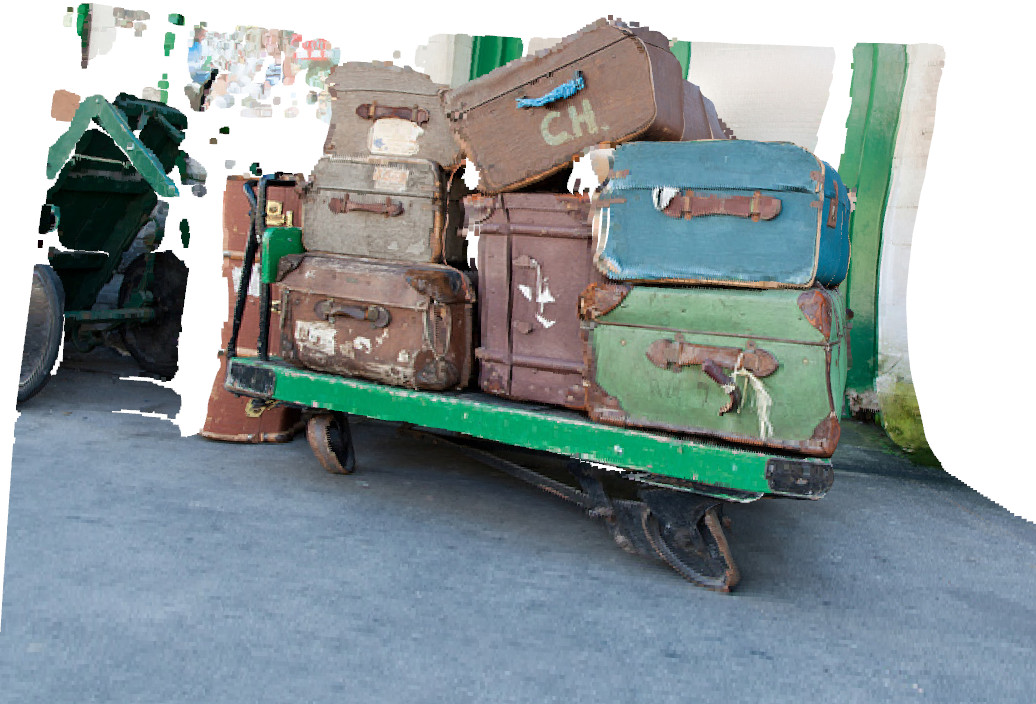} &
    \includegraphics[height=\teaserheight,trim={0cm 0 0.2cm 0}, clip]{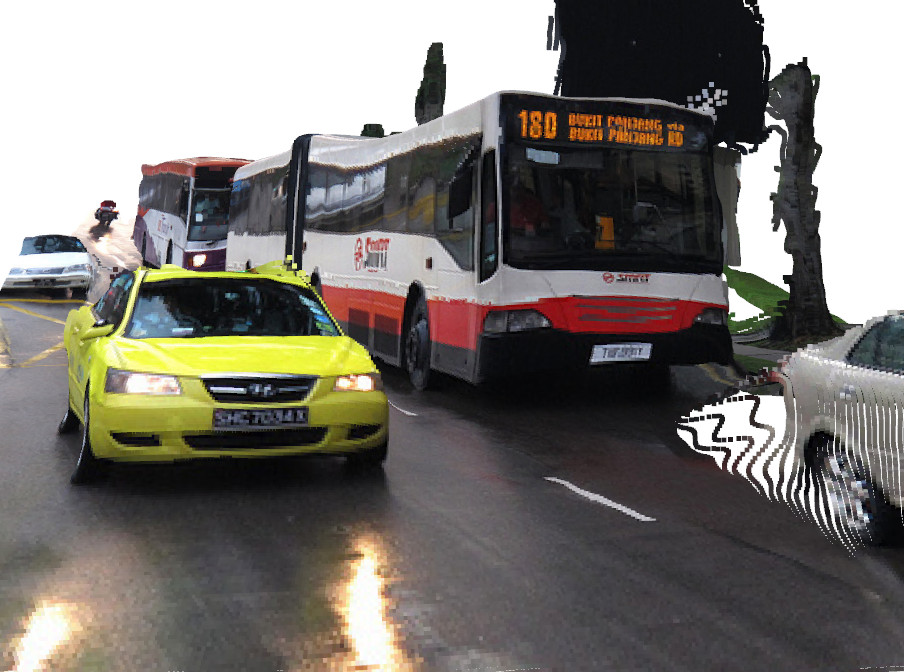}
	\end{tabular}
\caption{We show how to leverage training data from multiple, complementary sources for single-view depth estimation, in spite of
  varying and unknown depth range and scale. Our approach enables strong generalization across datasets. Top: input  
  images.
  Middle: inverse depth maps predicted by the presented approach.
  Bottom: corresponding point clouds rendered from a novel view-point. Point clouds
  rendered via Open3D~\cite{Zhou2018}. Input images from the Microsoft COCO
  dataset~\cite{Lin2014}, which was not seen during training.}
  \label{fig:teaser}
\end{figure*}

To learn models that are effective across a variety of scenarios, we need training data that is equally varied and captures the diversity of the visual world. The key challenge is to acquire such data at sufficient scale.
Sensors that provide dense ground-truth depth in dynamic scenes, such as structured light or time-of-flight, have limited range and operating conditions~\cite{KhoshelhamElberink2012,Hansard2013,Fankhauser2015}.
Laser scanners are expensive and can only provide sparse depth measurements when the scene is in motion. Stereo
cameras are a promising source of data~\cite{Garg2016,Godard2017}, but collecting suitable stereo images in diverse environments at scale remains a challenge.
Structure-from-motion (SfM) reconstruction has been used to construct training data for monocular depth estimation across a
variety of scenes~\cite{LiSnavely2018}, but the result does not include independently moving objects and is incomplete due
to the limitations of multi-view matching.
On the whole, none of the existing datasets is sufficiently rich to support the training of a model
that works robustly on real images of diverse scenes.
At present, we are faced with multiple datasets that may usefully complement each other, but are individually biased and incomplete.

In this paper, we investigate ways to train robust monocular depth estimation models that are expected to perform across diverse environments. We develop novel loss functions that are invariant to the major sources of
incompatibility between datasets, including unknown and inconsistent scale and baselines. Our losses enable training on data
that was acquired with diverse sensing modalities such as stereo cameras (with potentially unknown calibration), laser scanners,
and structured light sensors.
We also quantify the value of a variety of existing datasets for monocular depth estimation and explore optimal
strategies for mixing datasets during training. In particular, we show that a principled approach based on multi-objective
optimization~\cite{Sener2018} leads to improved results compared to a naive mixing strategy. We further empirically
highlight the importance of high-capacity encoders, and show the unreasonable effectiveness of pretraining the encoder on
a large-scale auxiliary task.

Our extensive experiments, which cover approximately six GPU months of computation, show that a model trained on a rich and diverse set of images from different sources, with an appropriate training procedure, delivers state-of-the-art results across a variety of environments.
To demonstrate this, we use the experimental protocol of \emph{zero-shot cross-dataset transfer}. That is, we train a model on certain datasets and then test its performance on other datasets that were never seen during training. The intuition is that zero-shot cross-dataset performance is a more faithful proxy of ``real world'' performance than training and testing on subsets of a single data collection that largely exhibit the same biases~\cite{TorralbaEfros2011}.

In an evaluation across six different datasets, we outperform prior art both quantitatively and qualitatively, and set a
new state of the art for monocular depth estimation. Example results are shown in Figure~\ref{fig:teaser}.

\section{Related Work}
\label{sec:related}
Early work on monocular depth estimation used MRF-based formulations~\cite{Saxena2009}, simple geometric assumptions~\cite{Hoiem2005}, or non-parametric methods~\cite{Karsch2014}.
More recently, significant advances have been made by leveraging the expressive power of convolutional networks to directly regress scene depth from the input image~\cite{Eigen2014}. Various architectural innovations have been proposed to enhance prediction accuracy~\cite{Laina2016,Roy2016,Liu2015,Fu2018,Li18}.
These methods need ground-truth depth for training, which is commonly acquired using \mbox{RGB-D} cameras or LiDAR sensors.
Others leverage existing stereo matching methods to obtain ground truth for supervision~\cite{Guo2018,Luo2018}.
These methods tend to work well in the specific type of scenes used to train them, but do not generalize well to unconstrained scenes, due to the limited scale and diversity of the training data.

Garg \etal \cite{Garg2016} proposed to use calibrated stereo cameras for self-supervision. While this
significantly simplifies the acquisition of training data, it still does not lift the restriction to a very specific data regime.
Since then, various approaches leverage self-supervision, but they either require stereo images
\cite{Godard2017,Zhan2018,Godard2019} or exploit apparent motion \cite{Zhou2017,Mahjourian2018,Casser2019,Godard2019}, and
are thus difficult to apply to dynamic scenes.

We argue that high-capacity deep models for monocular depth estimation can in principle operate on a fairly wide and
unconstrained range of scenes. What limits their performance is the lack of large-scale, dense ground truth that spans such a wide range of conditions.
Commonly used datasets feature homogeneous scene layouts, such as street scenes in a specific geographic
region~\cite{Geiger2012,MenzeGeiger2015,Saxena2009} or indoor environments~\cite{Silberman2012}. We note in particular that
these datasets show only a small number of dynamic objects. Models that are trained on data with such strong biases are prone to fail in less constrained environments.

\begin{table*}[!t]
  \caption{Datasets used in our work. Top: Our training sets. Bottom: Our test sets. No single real-world dataset features a large number of diverse scenes with dense and accurate ground truth.}
  \label{tab:datasets}
  \centering
  \scalebox{1.00}{
  \ra{1.05}
\begin{tabular}{@{}l@{\hspace{0mm}}c@{\hspace{2mm}}c@{\hspace{2mm}}c@{\hspace{2mm}}c@{\hspace{2mm}}c@{\hspace{2mm}}c@{\hspace{2mm}}c@{\hspace{2mm}}c@{\hspace{2mm}}c@{\hspace{2mm}}c@{\hspace{2mm}}c@{}}
	\toprule
	Dataset                                          & Indoor & Outdoor & Dynamic & Video  & Dense  & Accuracy & Diversity & Annotation   & Depth & \# Images      \\

\midrule
      DIML Indoor~\cite{Kim2018}                        & \YesV  & &              & \YesV &  \YesV & Medium & Medium & RGB-D & \bf{Metric} & 220K \\%/1500 \\
      MegaDepth~\cite{LiSnavely2018}                   &        & \YesV   & (\YesV) &             & (\YesV)  &  Medium  &   Medium  & SfM      & No scale    & 130K\\ %
      ReDWeb~\cite{Xian2018}                           & \YesV  & \YesV   &  \YesV  &             & \YesV  &  Medium  &   \bf{High}    & Stereo       & No scale \& shift & 3600 \\ %
      WSVD~\cite{Wang2019}                            &  \YesV  & \YesV   &  \YesV  & \YesV &  \YesV & Medium & \bf{High} & Stereo & No scale \& shift & 1.5M \\ %
      3D Movies                                       & \YesV  & \YesV   &  \YesV  & \YesV &  \YesV  &  Medium  &   \bf{High}    & Stereo  & No scale \& shift     & 75K \\ %

\midrule
        DIW~\cite{Chen2016}                              & \YesV  & \YesV   &  \YesV  &       &        &  Low     &   \bf{High}    & User clicks & Ordinal pair & 496K \\ %
        ETH3D~\cite{Schoeps2017}                               & \YesV  & \YesV   &         &        & \YesV  &  \bf{High}    &   Low  & Laser       &  \bf{Metric} & 454               \\ %
        Sintel~\cite{Butler2012}                          & \YesV  & \YesV   &  \YesV  & \YesV  & \YesV  &  \bf{High}    &   Medium    & Synthetic   &  (Metric) & 1064               \\ %
	KITTI~\cite{Geiger2012, MenzeGeiger2015}  &        & \YesV   &  (\YesV)  & \YesV &   (\YesV)    &  Medium    &   Low     & Laser/Stereo        &  \bf{Metric} & 93K \\ %
	NYUDv2~\cite{Silberman2012}                      & \YesV  &         &   (\YesV)      & \YesV   & \YesV  &  Medium       &   Low     & RGB-D       &  \bf{Metric} & 407K \\ %
	TUM-RGBD~\cite{Sturm2012}                     & \YesV &             & (\YesV) & \YesV & \YesV & Medium & Low & RGB-D &  \bf{Metric} & 80K \\ %

      \bottomrule
	\end{tabular}
    }
\end{table*}

Efforts have been made to create more diverse datasets.
Chen \etal~\cite{Chen2016} used crowd-sourcing to sparsely annotate ordinal relations in images collected from the web.
Xian \etal~\cite{Xian2018} collected a stereo dataset from the web and used off-the-shelf tools to extract dense
ground-truth disparity; while this dataset is fairly diverse, it only contains 3,600 images.
Li and Snavely~\cite{LiSnavely2018} used SfM and multi-view stereo (MVS) to reconstruct many (predominantly static) 3D scenes for supervision. Li~\etal \cite{Li2019} used SfM and MVS to construct a dataset from videos of people imitating mannequins (\ie\ they are frozen in action while the camera moves through the scene).
Chen \etal~\cite{Chen2019} propose an approach to automatically assess the quality of sparse SfM reconstructions in order to construct a large dataset.
Wang \etal~\cite{Wang2019} build a large dataset from stereo videos sourced from the web, while Cho \etal~\cite{Cho2019}
collect a dataset of outdoor scenes with handheld stereo cameras. Gordon \etal~\cite{Gordon2019} estimate the intrinsic parameters of YouTube videos in order to leverage them for training.
Large-scale datasets that were collected from the Internet~\cite{Li2019,Wang2019} require a large amount of pre- and post-processing.
Due to copyright restrictions, they often only provide links to videos, which
frequently become unavailable. This makes reproducing these datasets challenging.

To the best of our knowledge, the controlled mixing of multiple data sources has not been explored before in this context.
Ummenhofer \etal~\cite{Ummenhofer2017} presented a model for two-view structure and motion estimation and trained it on a dataset of (static) scenes that is the union of multiple smaller datasets.
However, they did not consider strategies for optimal mixing, or study the impact of combining multiple datasets.
Similarly, Facil \etal~\cite{Facil2019} used multiple datasets with a naive mixing strategy for learning monocular depth
with known camera intrinsics. Their test data is very similar to half of their training
collection, namely RGB-D recordings of indoor scenes. %

\section{Existing Datasets}
\label{sec:dataset}
Various datasets have been proposed that are suitable for monocular depth estimation, \ie\ they consist of RGB images with corresponding depth annotation of some form~\cite{Saxena2009,Butler2012,Geiger2012,Silberman2012,Sturm2012,MenzeGeiger2015,Song2015,Cordts2016,Chen2016,Dai2017,
Knapitsch2017,Schoeps2017,LiSnavely2018,Kim2018,Xian2018,Cho2019,Li2019,Vasiljevic2019,Wang2019}.
Datasets differ in captured environments and objects (indoor/outdoor scenes, dynamic objects), type of depth annotation
(sparse/dense, absolute/relative depth), accuracy (laser, time-of-flight, SfM, stereo, human annotation, synthetic data), image
quality and camera settings, as well as dataset size.

Each single dataset comes with its own characteristics and has its own biases and
problems~\cite{TorralbaEfros2011}. High-accuracy data is hard to acquire at scale and problematic for dynamic objects~\cite{Knapitsch2017,Schoeps2017}, whereas large data collections from Internet sources come with limited image quality and depth accuracy as well as unknown camera parameters~\cite{Chen2016,Wang2019}.
Training on a single dataset leads to good performance on the corresponding test split of the same dataset (same camera parameters, depth annotation, environment), but may have limited generalization capabilities to unseen data with different characteristics.
Instead, we propose to train on a collection of datasets, and demonstrate that this approach leads to strongly
enhanced generalization by testing on diverse datasets that were not seen during training.
We list our training and test datasets, together with their individual characteristics, in Table \ref{tab:datasets}.

\mypara{Training datasets.}
We experiment with five existing and complementary datasets for training.
ReDWeb \cite{Xian2018} (RW) is a small, heavily curated dataset that features diverse and dynamic scenes with ground truth that was acquired with a relatively large stereo baseline.
MegaDepth~\cite{LiSnavely2018} (MD) is much larger, but shows predominantly static scenes.
The ground truth is usually more accurate in background regions since wide-baseline multi-view stereo reconstruction was used for acquisition.
WSVD~\cite{Wang2019} (WS) consists of stereo videos obtained from the web and features diverse and dynamic scenes. This
dataset is only available as a collection of links to the stereo videos. No ground truth is provided. We thus recreate the
ground truth according to the procedure outlined by the original authors.
DIML Indoor \cite{Kim2018} (DL) is an RGB-D dataset of predominantly static indoor scenes, captured with a Kinect v2.

\mypara{Test datasets.}
To benchmark the generalization performance of monocular depth estimation models, we chose six datasets based
on diversity and accuracy of their ground truth. DIW \cite{Chen2016} is highly diverse but provides ground truth only in
the form of sparse ordinal relations. ETH3D \cite{Schoeps2017} features highly accurate laser-scanned ground truth on
static scenes. Sintel \cite{Butler2012} features perfect ground truth for synthetic scenes.
KITTI~\cite{MenzeGeiger2015} and NYU~\cite{Silberman2012} are commonly used datasets with characteristic biases.
For the TUM dataset~\cite{Sturm2012}, we use the \textit{dynamic} subset that features humans in indoor environments~\cite{Li2019}.
Note that we never fine-tune models on any of these datasets. We refer to this experimental procedure as \emph{zero-shot
  cross-dataset transfer}.

\section{3D Movies}
\label{sec:movies}
\begin{table}[htb!]
\caption{List of films and the number of extracted frames in the 3D Movies dataset after automatic processing.}
  \centering
  \begin{tabular}{lr}
      \toprule
      Movie title & \# frames \\
      \midrule
      \textbf{Training set} & \textbf{75074} \\
Battle of the Year (2013) & 4821 \\ %
Billy Lynn's Long Halftime Walk (2016) & 4178 \\ %
Drive Angry (2011) & 328 \\
Exodus: Gods and Kings (2014) & 8063 \\
Final Destination 5 (2011) & 1437 \\
A very Harold \& Kumar 3D Christmas (2011) & 3690 \\
Hellbenders (2012) &120 \\
The Hobbit: An Unexpected Journey (2012) & 8874 \\
Hugo (2011) & 3189 \\
The Three Musketeers (2011) & 5028 \\
Nurse 3D (2013) & 492 \\
Pina (2011) & 1215 \\ %
Dawn of the Planet of the Apes (2014) & 5571 \\
The Amazing Spider-Man (2012) & 5618 \\
Step Up 3D (2010) & 509 \\
Step Up: All In (2014) & 2187 \\
Transformers: Age of Extinction (2014) & 8740 \\
Le Dernier Loup / Wolf Totem (2015) & 4843 \\
X-Men: Days of Future Past (2014) & 6171 \\
\midrule
      \textbf{Validation set} & \textbf{3058} \\
The Great Gatsby (2013) & 1815 \\
Step Up: Miami Heat / Revolution (2012) & 1243 \\ %
\midrule
      \textbf{Test set} & \textbf{788} \\
Doctor Who - The Day of the Doctor (2013) & 508 \\
StreetDance 2 (2012) & 280 \\
    \bottomrule
	\end{tabular}
\label{tab:movie_list}
\end{table}

\begin{figure*}[t!]
  \centering
  \newcommand{\imgSize}{0.245\textwidth}
 	\begin{tabular}{@{}c@{\hspace{1mm}}c@{\hspace{1mm}}c@{\hspace{1mm}}c@{}}

  \includegraphics[width=\imgSize]{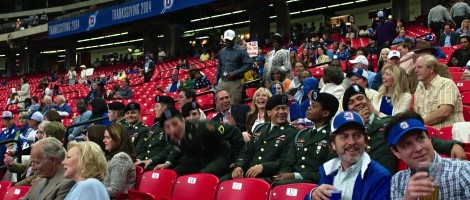} &
  \includegraphics[width=\imgSize]{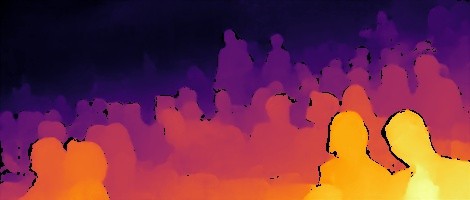} &
    \includegraphics[width=\imgSize]{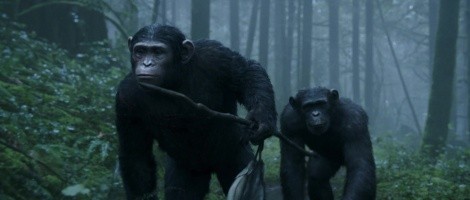} &
  \includegraphics[width=\imgSize]{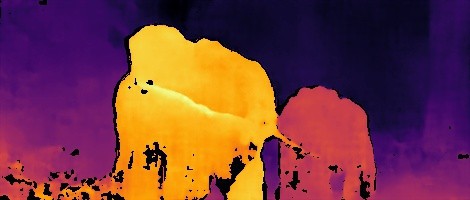}
  \vspace{0mm}\\
    \includegraphics[width=\imgSize]{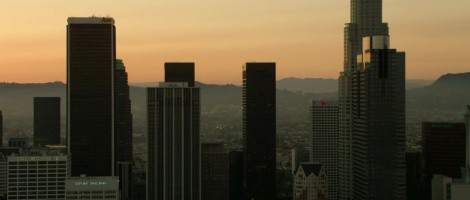} &
  \includegraphics[width=\imgSize]{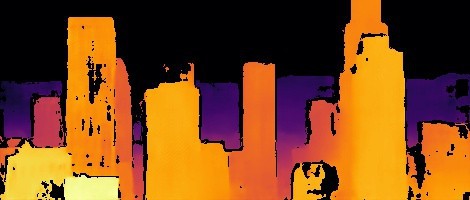}  &
  \includegraphics[width=\imgSize]{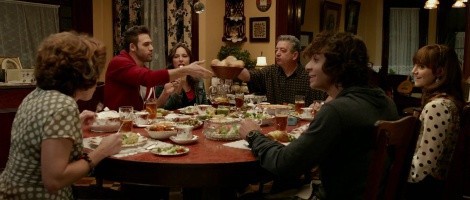} &
  \includegraphics[width=\imgSize]{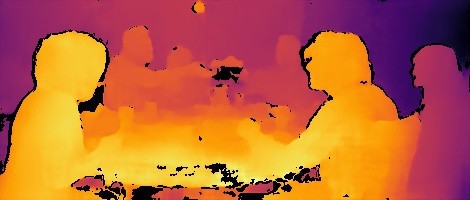}
  \vspace{0mm}\\
  \includegraphics[width=\imgSize]{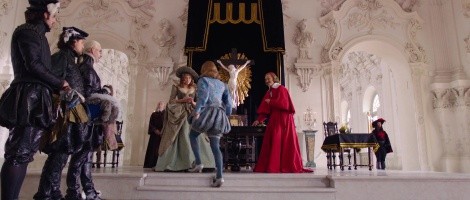} &
  \includegraphics[width=\imgSize]{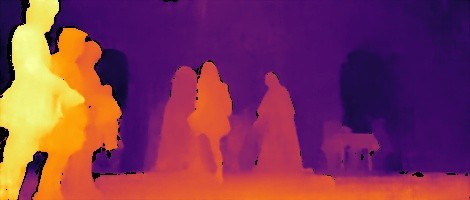}  &

    \includegraphics[width=\imgSize]{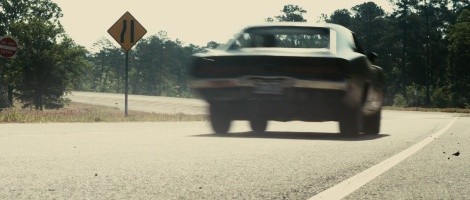} &
  \includegraphics[width=\imgSize]{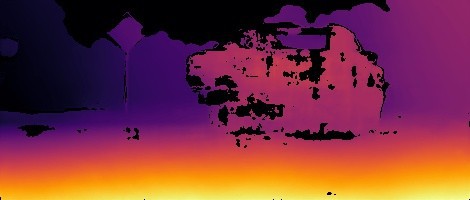}
  \vspace{0mm}\\
    \includegraphics[width=\imgSize]{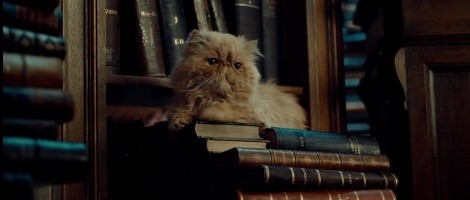} &
  \includegraphics[width=\imgSize]{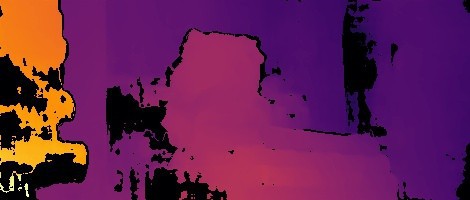} &
  \includegraphics[width=\imgSize]{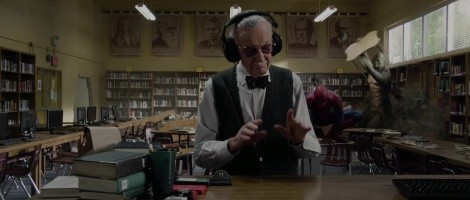} &
  \includegraphics[width=\imgSize]{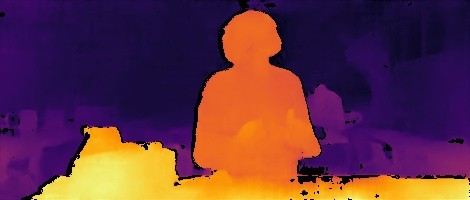}
  \vspace{0mm}\\

  \includegraphics[width=\imgSize]{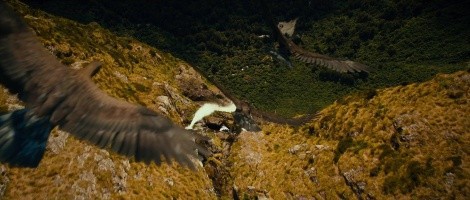} &
 \includegraphics[width=\imgSize]{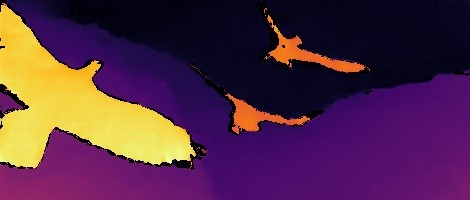} &
    \includegraphics[width=\imgSize]{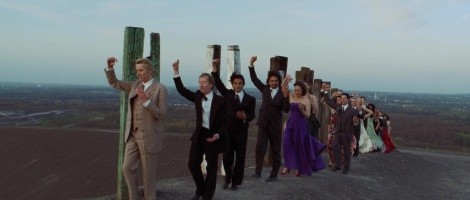} &
  \includegraphics[width=\imgSize]{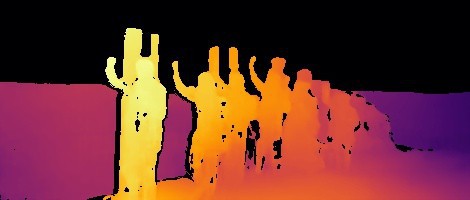}

   \vspace{0mm}\\
  \includegraphics[width=\imgSize]{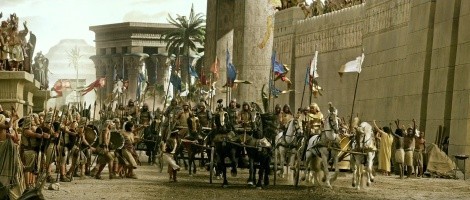}&
  \includegraphics[width=\imgSize]{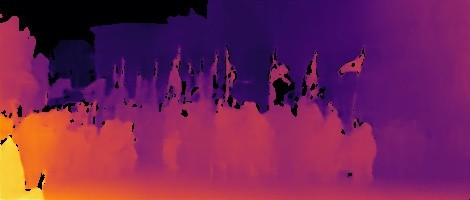} &
  \includegraphics[width=\imgSize]{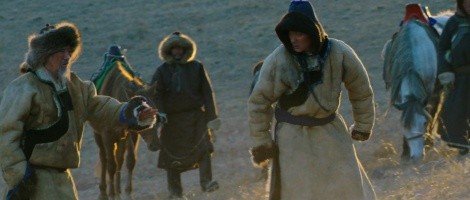} &
  \includegraphics[width=\imgSize]{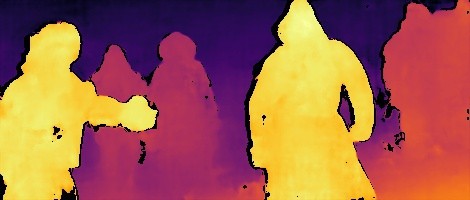}
  \vspace{0mm}

 	\end{tabular}
    \caption{Sample images from the 3D Movies dataset. We show images from some of the films in the training set together with their inverse depth maps. Sky regions and invalid pixels are masked out. Each image is taken from a different film. 3D movies provide a massive source of diverse data.}
  \label{fig:dataset}
\end{figure*}

To complement the existing datasets we propose a new data source: 3D movies (MV).
3D movies feature high-quality video frames in a variety of dynamic environments that range from human-centric imagery in story- and dialogue-driven Hollywood films to nature scenes with landscapes and animals in documentary features. While the data does not provide metric depth, we can use stereo matching to obtain relative depth (similar to RW and WS).
Our driving motivation is the scale and diversity of the data. 3D movies provide the largest known source of stereo pairs
that were captured in carefully controlled conditions. This offers the possibility of tapping into millions of high-quality
images from an ever-growing library of content. We note that 3D movies have been used in related tasks in
isolation~\cite{Hadfield2017,Xie2016}. We will show that their full potential is unlocked by combining them with other,
complementary data sources. In contrast to similar data collections in the wild~\cite{Xian2018,Wang2019,Li2019}, no manual
filtering of problematic content was required with this data source. Hence, the dataset can easily be extended or adapted to specific needs (\eg\ focus on dancing humans or nature documentaries).

\mypara{Challenges.}
Movie data comes with its own challenges and imperfections.
The primary objective when producing stereoscopic film is providing a visually pleasing viewing experience while avoiding discomfort for the viewer~\cite{DevernayBeardsley2010}.
This means that the disparity range for any given scene (also known as the depth budget) is limited and depends on both artistic and psychophysical considerations.
For example, disparity ranges are often increased in the beginning and the end of a movie, in order to induce a very noticeable stereoscopic effect for a short time. Depth budgets in the middle may be lower to allow for more comfortable viewing.
Stereographers thus adjust their depth budget depending on the content, transitions, and even the rhythm of scenes\changed{~\cite{Neuman2009}}.

In consequence, focal lengths, baseline, and convergence angle between the cameras of the stereo rig are unknown and vary between scenes even within a single film.
Furthermore, in contrast to image pairs obtained directly from a standard stereo camera, stereo pairs in movies usually contain both positive and negative disparities to allow objects to be perceived either in front of or behind the screen.
Additionally, the depth that corresponds to the screen is scene-dependent and is often modified in post-production by shifting the image pairs.
We describe data extraction and training procedures that address these challenges.

\mypara{Movie selection and preprocessing.}
We selected a diverse set of 23 movies. The selection was based on the following considerations:
1)~We only selected movies that were shot using a physical stereo camera. (Some 3D films are shot with a monocular camera and the stereoscopic effect is added in post-production by artists.)
2)~We tried to balance realism and diversity.
3)~We only selected movies that are available in Blu-ray format and thus allow extraction of high-resolution images.

We extract stereo image pairs at 1920x1080 resolution and 24 frames per second (fps).
Movies have varying aspect ratios, resulting in black bars on the top and bottom of the frame, and some movies have thin black bars along frame boundaries due to post-production. We thus center-crop all frames to 1880x800 pixels.
We use the chapter information (Blu-ray meta-data) to split each movie into individual chapters.
We drop the first and last chapters since they usually include the introduction and credits.

We use the scene detection tool of FFmpeg~\cite{ffmpeg2018} with a threshold of 0.1 to extract individual clips.
We discard clips that are shorter than one second to filter out chaotic action scenes and highly correlated
clips that rapidly switch between protagonists during dialogues.
To balance scene diversity, we sample the first 24 frames of each clip
and additionally sample 24 frames every four seconds for longer clips.
Since multiple frames are part of the same clip, the complete dataset is highly correlated.
Hence, we further subsample the training set at 4 fps and the test and validation sets at 1 fps.

\mypara{Disparity extraction.}
The extracted image pairs can be used to estimate disparity maps using stereo matching.
Unfortunately, state-of-the-art stereo matchers perform poorly when applied to movie data,
since the matchers were designed and trained to match only over positive disparity ranges. This assumption is appropriate for the rectified output of a standard stereo camera, but not to image pairs extracted from stereoscopic film.
Moreover, disparity ranges encountered in 3D movies are usually smaller than ranges that are common in standard stereo setups due to the limited depth budget.

To alleviate these problems, we apply a modern optical flow algorithm~\cite{Sun2018} to the stereo pairs. We retain the horizontal component of the flow as a proxy for disparity.
Optical flow algorithms naturally handle both positive and negative disparities and usually perform well for displacements of moderate size.
For each stereo pair we use the left camera as the reference and extract the optical flow from the left to the right image and vice versa.
We perform a left-right consistency check and mark pixels with a disparity difference of more than 2 pixels as invalid.
We automatically filter out frames of bad disparity quality following the guidelines of Wang \etal \cite{Wang2019}: frames are rejected if more than 10\% of all pixels have a vertical disparity \textgreater 2 pixels, the horizontal disparity range is \textless 10 pixels, or the percentage of pixels passing the left-right consistency check is \textless 70\%.
In a final step, we detect pixels that belong to sky regions using a pre-trained semantic segmentation model \cite{RotaBulo2018} and set their disparity to the minimum disparity in the image.

The complete list of selected movies together with the number of frames that remain after filtering with the automatic
cleaning pipeline is shown in Table \ref{tab:movie_list}. Note that discrepancies in the number of extracted frames per movie occur due to varying runtimes as well as varying disparity quality.
We use frames from 19 movies for training and set aside two movies for validation and two movies for testing, respectively.
Example frames from the resulting dataset are shown in Figure~\ref{fig:dataset}.

\section{Training on Diverse Data}
\label{sec:method}
Training models for monocular depth estimation on diverse datasets presents a challenge because the ground truth comes in different forms (see Table \ref{tab:datasets}).
It may be in the form of absolute depth (from laser-based measurements or stereo cameras with known calibration), depth up to an unknown scale (from SfM), or disparity maps (from stereo cameras with unknown calibration).
The main requirement for a sensible training scheme is to carry out computations in an
appropriate output space that is compatible with all ground-truth representations and is numerically well-behaved.
We further need to design a loss function that is flexible enough to handle diverse sources of data while making optimal use of all available information.

We identify three major challenges.
1)~Inherently different representations of depth: direct vs.\ inverse depth representations.
2)~Scale ambiguity: for some data sources, depth is only given up to an unknown scale.
3)~Shift ambiguity: some datasets provide disparity only up to an unknown scale and global disparity shift that is a function of the
unknown baseline and a horizontal shift of the principal points due to post-processing~\cite{Wang2019}. %

\mypara{Scale- and shift-invariant losses.}
We propose to perform prediction in disparity space (inverse depth up to scale and shift) together with a family of scale- and shift-invariant dense losses to handle the aforementioned ambiguities.
Let $M$ denote the number of pixels in an image with valid ground truth and let $\btheta$ be the parameters of the prediction model.
Let ${\dd = \dd(\btheta) \in \Re^{M}}$ be a disparity prediction and let ${\dd^* \in \Re^{M}}$ be the corresponding ground-truth disparity.
Individual pixels are indexed by subscripts.

We define the scale- and shift-invariant loss for a single sample as
\begin{equation}
\lL_{ssi}(\hat \dd, \hat \dd^*) = \frac{1}{2M} \sum_{i=1}^M \rho \left( \hat \dd_i - \hat \dd_i^* \right),
\label{eq:ssi}
\end{equation}
where $\hat \dd$ and $\hat \dd^*$ are scaled and shifted versions of the predictions and ground truth, and $\rho$
defines the specific type of loss function.

Let $s: \Re^{M} \rightarrow \Re_+$ and $t: \Re^{M} \rightarrow \Re$ denote estimators of the scale and translation. To
define a meaningful scale- and shift-invariant loss, a sensible requirement is that prediction and ground truth should be
appropriately aligned with respect to their scale and shift, \ie\ we need to ensure that  $s(\hat\dd) \approx s(\hat\dd^*)$ and $t(\hat\dd) \approx t(\hat\dd^*)$.
We propose two different strategies for performing this alignment.%

The first approach aligns the prediction to the ground truth based on a least-squares criterion:
\begin{align}
  (s, t) &= \arg \min_{s,t} \sum_{i=1}^M \left( s\dd_i + t - \dd^*_i\right)^2\nonumber \,,\\
  \hat \dd &= s \dd + t,\quad\hat \dd^* = \dd^*,
  \label{eq:ls_align}
\end{align}
where $\hat \dd$ and $\hat \dd^*$ are the aligned prediction and ground truth, respectively.
The factors $s$ and $t$ can be efficiently determined in closed form by rewriting \eqref{eq:ls_align} as a
standard least-squares problem:
Let $\vec \dd_i = (\dd_i, 1)\trans$ and $\hh = (s, t)\trans$, then we can
rewrite the objective as
\begin{align}
  \hh^{opt} = \arg \min_\hh \sum_{i=1}^M \left(\vec \dd_i\trans \hh  - \dd_i^*\right)^2 ,
\label{eq:ls_align_vectorized}
\end{align}
which has the closed-form solution
\begin{align}
 \hh^{opt} = \left(\sum_{i=1}^M \vec \dd_i \vec \dd_i\trans\right)^{-1} \left(\sum_{i=1}^M \vec \dd_i \dd_i^*\right).
\end{align}
We set $\rho(x) = \rho_{mse}(x) = x^2$ to define the scale- and shift-invariant mean-squared error (MSE). We denote this loss as $\lL_{ssimse}$.

The MSE is not robust to the presence of outliers. %
Since all existing large-scale datasets only provide imperfect ground truth, we conjecture that a robust
loss function can improve training.
We thus define alternative, robust loss functions based on robust estimators of scale and shift:
\begin{align}
 t(\dd) = \mathrm{median}(\dd),\quad s(\dd) = \frac{1}{M} \sum_{i=1}^M |\dd - t(\dd)|.
\end{align}
We align both the prediction and the ground truth to have zero translation and unit scale:
\begin{align}
    \hat \dd &= \frac{\dd - t(\dd)}{s(\dd)},\quad\hat \dd^* = \frac{\dd^* - t(\dd^*)}{s(\dd^*)}.
\end{align}
We define two robust losses. The first, which we denote as  $\lL_{ssimae}$, measures the absolute deviations
${\rho_{mae}(x) = |x|}$. We define the second robust loss by trimming the 20\% largest residuals in
every image, irrespective of their magnitude:
\begin{align}
  \lL_{ssitrim}(\hat \dd, \hat \dd^*) = \frac{1}{2M} \sum_{j=1}^{U_m} \rho_{mae} \left( \hat \dd_j - \hat \dd_j^* \right),
\end{align}
with $|\hat \dd_j - \hat \dd_j^* | \leq | \hat \dd_{j+1} - \hat \dd_{j+1}^* |$ and $U_m = 0.8M$ \changed{(set empirically based on experiments on the ReDWeb dataset)}.
Note that this is in contrast to commonly used M-estimators, where the influence of large residuals is merely down-weighted. Our reasoning for trimming is that outliers in the ground truth should never influence training.

\mypara{Related loss functions.}
The importance of accounting for unknown or varying scale in the training of
monocular depth estimation models has been recognized early. Eigen \etal~\cite{Eigen2014} proposed a scale-invariant
loss in log-depth space. Their loss can be written as
\begin{align}
  \resizebox{.9\hsize}{!}{$\displaystyle{
  \lL_{silog}(\zz, \zz^*) = \min_{s} \frac{1}{2M} \sum_{i=1}^M \big(\log(e^s\zz_i) - \log(\zz_i^*) \big)^2,
  }$}
  \label{eq:eigen_loss}
\end{align}
where $\zz_i\! =\! \dd_i^{-1}$ and $\zz_i^*\! =\! (\dd_i^*)^{-1}$ are depths up to unknown scale.
Both \eqref{eq:eigen_loss} and $\lL_{ssimse}$ account for the unknown scale of the predictions, but only $\lL_{ssimse}$ accounts for an unknown global disparity shift. Moreover, the losses
are evaluated on different depth representations. Our loss is defined in disparity space, which is
numerically stable and  compatible with common representations of relative depth.

Chen \etal~\cite{Chen2016} proposed a generally applicable loss for relative depth estimation based on ordinal relations:
\begin{align}
  \resizebox{.9\hsize}{!}{$\displaystyle{
  \rho_{ord}(\zz_i - \zz_j) =
  \begin{cases}
    \log(1+\exp(-(\zz_{i}-\zz_{j})l_{ij}), & l_{ij} \ne 0\\
    (\zz_{i}-\zz_{j})^2, & l_{ij} = 0,
  \end{cases}
  }$}
\label{eq:ordloss2}
\end{align}
where $l_{ij} \! \in \! \left\{-1, 0, 1\right\}$
encodes the ground-truth ordinal relation of point pairs. This encourages pushing
points as far apart as possible when $l_{ij} \ne 0$ and pulling them to the same depth when $l_{ij} = 0$.
Xian \etal~\cite{Xian2018} suggest to sparsely evaluate this loss by randomly sampling point pairs from the dense ground truth. %
In contrast, our proposed losses take all available data into account.

Recently, Wang \etal~\cite{Wang2019} proposed the normalized multiscale gradient (NMG) loss.
To achieve shift invariance in addition to scale invariance in disparity space, they evaluate the gradient difference between ground-truth and rescaled estimates at multiple scales $k$:
\begin{align}
  \resizebox{.9\hsize}{!}{$\displaystyle{
  \lL_{nmg}(\dd, \dd^*) =  \sum_{k=1}^K  \sum_{i=1}^M | s \nabla_x^k \dd - \nabla_x^k \dd^* | + | s \nabla_y^k \dd - \nabla_y^k \dd^* |.
  }$}
  \label{eq:nmg_loss}
\end{align}
In contrast, our losses are evaluated directly on the ground-truth disparity values, while also accounting for unknown scale
and shift.
While both the ordinal loss and NMG can, conceptually, be applied to arbitrary depth representations and are thus suited for mixing diverse datasets, we will show that our scale- and shift-invariant loss variants lead to consistently better performance.

\mypara{Final loss.}
To define the complete loss, we adapt the multi-scale, scale-invariant gradient matching term \cite{LiSnavely2018} to the disparity space. This term
biases discontinuities to be sharp and to coincide with discontinuities in the ground truth.
We define the gradient matching term as
\begin{align}
  \lL_{reg}(\hat \dd, \hat \dd^*) = \frac{1}{M} \sum_{k=1}^K \sum_{i=1}^{M} \big(|\nabla_x R_i^k|+|\nabla_y R_i^k|\big),
  \label{eq:lgrad}
\end{align}
where $R_i = \hat \dd_i- \hat \dd_i^*$, and $R^k$ denotes the difference of disparity maps at scale $k$.
We use $K=4$ scale levels, halving the image resolution at each level. Note that this term is similar to $\lL_{nmg}$, but
with different approaches to compute the scaling $s$.

Our final loss for a training set $l$ is
\begin{align}
  \resizebox{.88\hsize}{!}{
    $\displaystyle{
      \lL_l = \frac{1}{N_l}\sum_{n=1}^{N_l} \lL_{ssi}\big(\hat \dd^n, (\hat \dd^*)^n\big) + \alpha \; \lL_{reg}\big(\hat \dd^n,
      (\hat \dd^*)^n\big),}$
  }
  \label{eq:total_loss}
\end{align}
where $N_l$ is the training set size and $\alpha$ is set to 0.5.

\mypara{Mixing strategies.}
While our loss and choice of prediction space enable mixing datasets, it is not immediately clear in what proportions different datasets should be integrated during training with a stochastic optimization algorithm.
We explore two different strategies in our experiments.

The first, naive strategy is to mix datasets in equal parts in each minibatch. For a minibatch of size $B$, we sample $B/L$ training samples from each dataset, where $L$ denotes the number of distinct datasets.
This strategy ensures that all datasets are represented equally in the effective training set, regardless of their individual size.

Our second strategy explores a more principled approach, where we adapt a recent procedure for Pareto-optimal multi-task learning to our setting \cite{Sener2018}.
We define learning on each dataset as a separate task and seek an approximate Pareto optimum over datasets (\ie\ a solution where the loss cannot be decreased on any training set without increasing it for at least one of the others).
Formally, we use the algorithm presented in \cite{Sener2018}  to minimize the multi-objective optimization criterion
\begin{align}
  \min_{\btheta} \big(\lL_1(\btheta), \dots, \lL_L(\btheta)\big)\trans,
\end{align}
where model parameters $\btheta$ are shared across datasets.

\section{Experiments}
\label{sec:experiments}
\newcommand{\red}[1] {\color{Red}#1}
\newcommand{\green}[1] {\color{Green}#1}

\newcommand{\changedt}[1] {#1}
\newcommand{\changedg}[1] {\color{Green}#1}
\newcommand{\changedr}[1] {\color{Red}#1}
\def\Uline#1{#1\llap{\uline{\phantom{11.11}}}}

We start from the experimental setup of Xian \etal~\cite{Xian2018} and use their ResNet-based \cite{He2016} multi-scale architecture for single-image depth prediction.
We initialize the encoder with pretrained ImageNet~\cite{Deng2009} weights and initialize other layers randomly.
We use Adam \cite{Kingma2015} with a learning rate of $10^{-4}$ for randomly initialized layers and $10^{-5}$ for
pretrained layers, and set the exponential decay rate to $\beta_1 = 0.9$ and $\beta_2= 0.999$. Images are flipped horizontally with a 50\% chance, and randomly cropped and resized to $384\times384$ to augment the data and maintain the aspect ratio across different input images. \changed{No other augmentations are used.}
\removed{We pretrain the network for 300 epochs on RW to produce a baseline model comparable to [32].}

Subsequently, we perform ablation studies on the loss function and, since we conjecture that pretraining on ImageNet data
has significant influence on performance, also the encoder architecture. We use the best-performing pretrained model as the
starting point for our dataset mixing experiments.
We use a batch size of $8L$, \ie\ when mixing three datasets the batch size is $24$.
When comparing datasets of different sizes, the term epoch is not well-defined; we thus denote an epoch as
processing $72@000$ images, roughly the size of MD and MV, and train for $60$ epochs. We shift and scale the ground-truth disparity to the range $[0,1]$ for all datasets.

\begin{figure}[t]
	\centering
	\includegraphics[width=.95\columnwidth]{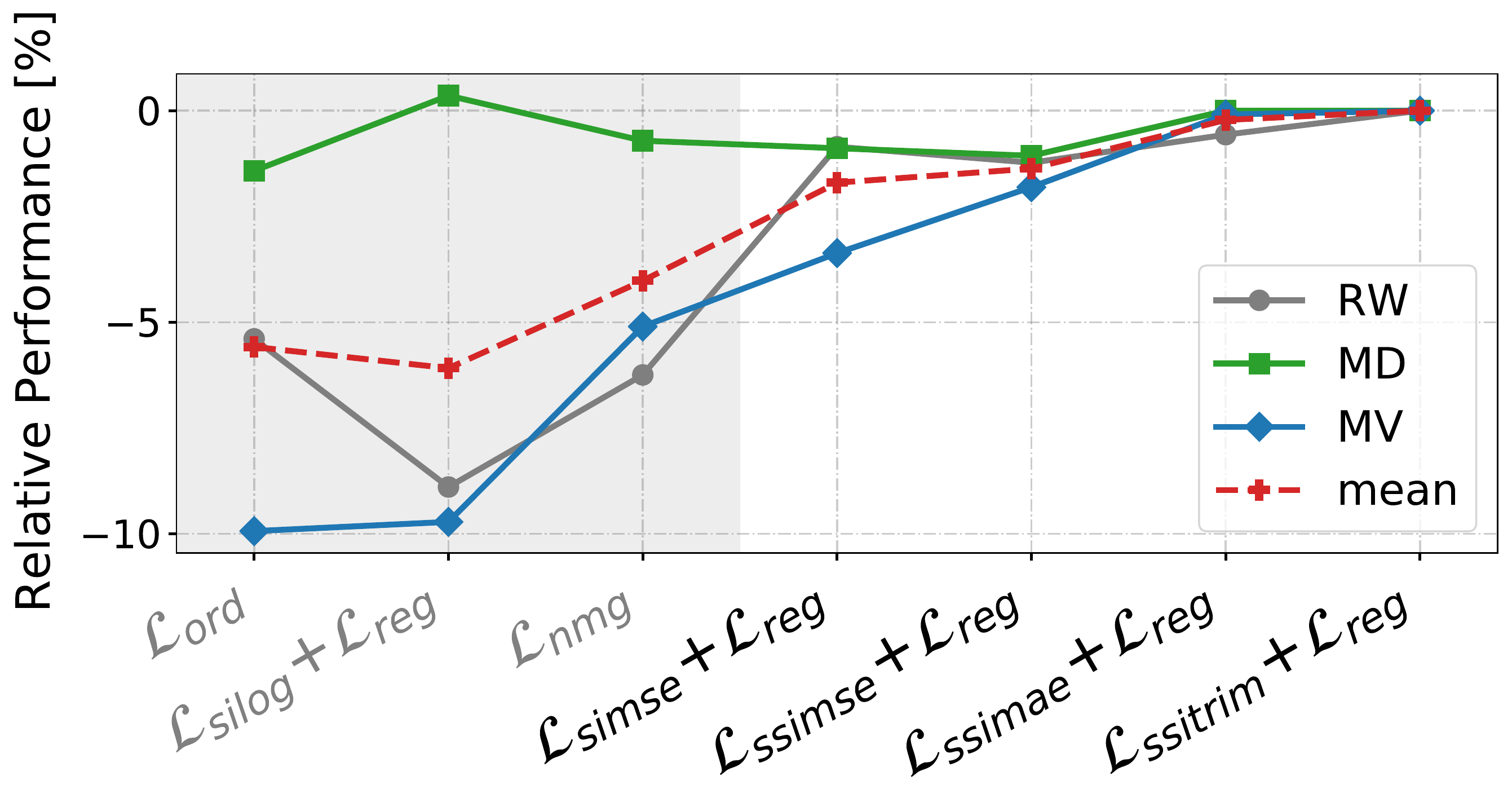}
	\caption{Relative performance of different loss functions (higher is better) with the best performing loss $\lL_{ssitrim}+\lL_{reg}$ used as reference. All our four proposed losses (white area) outperform current state-of-the-art losses (gray area).}
	\label{fig:loss}
  \end{figure}

\mypara{Test datasets and metrics.}
For ablation studies of loss and encoders, we use our held-out validation sets of RW (360 images), MD (2,963 images --
official validation set), and MV (3,058 images -- see Table \ref{tab:movie_list}).
For all training dataset mixing experiments and comparisons to the state of the art, we test on a collection of datasets
that were never seen during training: DIW, ETH3D, Sintel, KITTI, NYU, and TUM.
For DIW~\cite{Chen2016} we created a validation set of 10,000 images from the DIW training set for our ablation studies and used the official test set of 74,441 images when comparing to the state of the art.
For NYU we used the official test split (654 images). For KITTI we used the intersection of the official validation set for
depth estimation (with improved ground-truth depth \cite{Uhrig2017}) and the Eigen test
split~\cite{EigenFergus2015} (161 images). For ETH3D and Sintel we used the whole dataset for which ground truth is
available (454 and 1,064 images, respectively).
For the TUM dataset, we use the \textit{dynamic} subset that features humans in indoor environments~\cite{Li2019} (1,815 images).

For each dataset, we use a single metric that fits the ground truth in that dataset.
For DIW we use the Weighted Human Disagreement Rate \changed{(WHDR)}~\cite{Chen2016}.
For datasets that are based on relative depth, we measure the root mean squared error in disparity space (MV, RW, MD).
For datasets that provide accurate absolute depth \changed{(ETH3D, Sintel)}, we measure the mean absolute value of the relative error $(1/M) \sum_{i=1}^M \abs{z_i - z_i^*}/z_i^*$ in depth space \changed{(AbsRel)}.
Finally, we use the percentage of pixels with \changed{$\delta = \max(\frac{z_i}{z_i^*},\frac{z_i^*}{z_i})\!>\!1.25$} to evaluate models on
KITTI, NYU, and TUM~\cite{Eigen2014}.
Following~\cite{Godard2017}, we cap predictions at an appropriate maximum value for datasets that are evaluated in depth space.
For ETH3D, KITTI, NYU, and TUM, the depth cap was set to the maximum ground-truth depth value (72, 80, 10, and 10 meters, respectively).
For Sintel, we evaluate on areas with ground-truth depth below 72 meters and accordingly use a depth cap of 72 meters.
For all our models and baselines, we align predictions and ground truth in scale and shift \changed{for each image} before measuring errors.
We perform the alignment in inverse-depth space based on the least-squares criterion.
Since absolute numbers quickly become hard to interpret when evaluating on multiple datasets, we also present the relative
change in performance compared to an appropriate baseline method.

\begin{figure}[t]
\centering
      \includegraphics[width=.95\columnwidth]{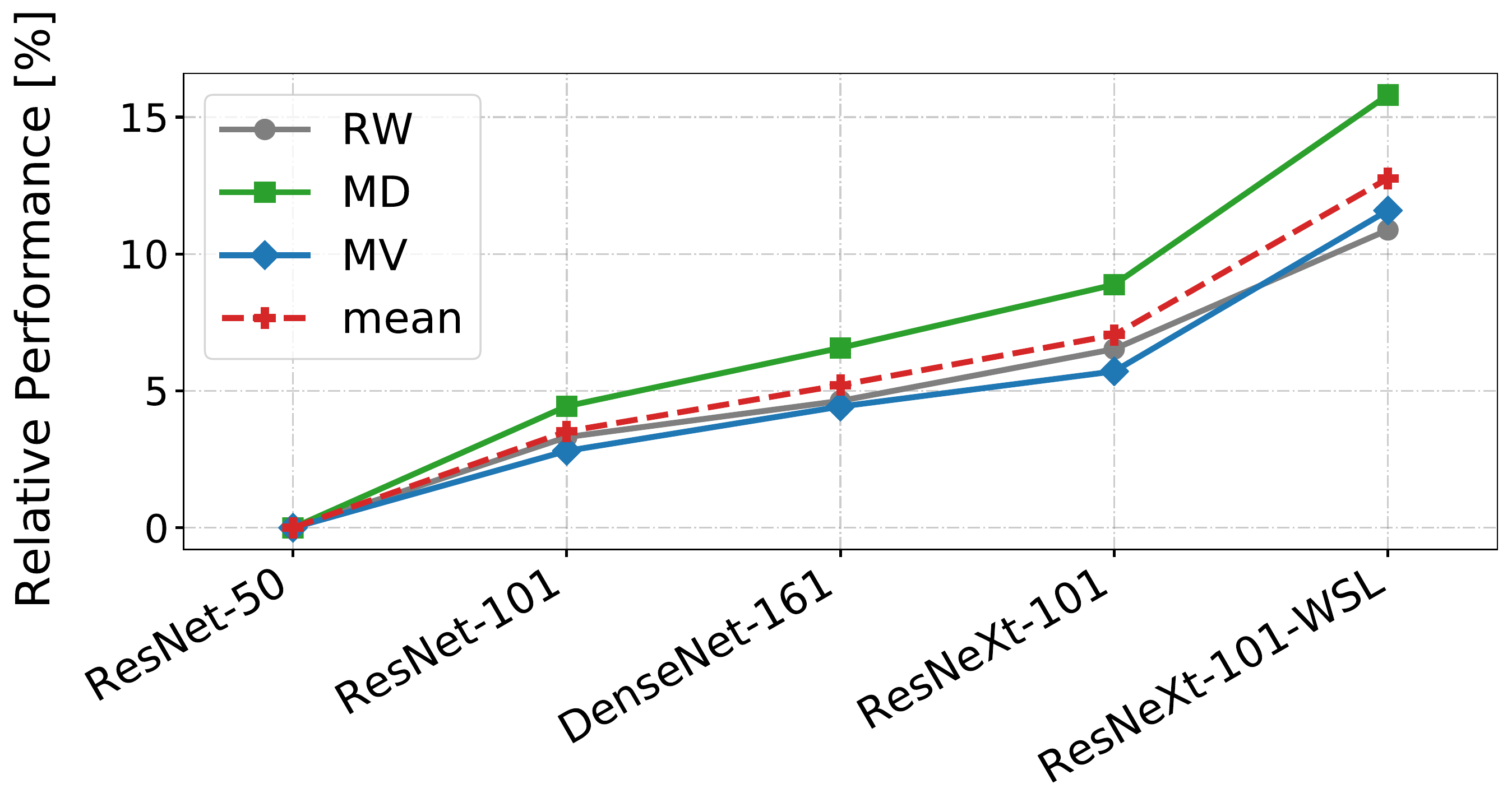}
      \caption{Relative performance of different encoders across datasets (higher is better).
        ImageNet performance of an encoder is predictive of its performance in monocular depth estimation.}
   \label{fig:encoder}
 \end{figure}

\mypara{Input resolution for evaluation.}
We resize test images so that the larger axis equals 384 pixels while the smaller axis is resized to a multiple of 32
pixels  (a constraint imposed by the encoder), while keeping an aspect ratio as close as possible to the original aspect ratio. Due to the wide
aspect ratio in KITTI this strategy would lead to very small input images. We thus resize the \textit{smaller} axis to be equal to
384 pixels on this dataset and adopt the same strategy otherwise to maintain the aspect ratio.

Most state-of-the-art methods that we compare to are specialized to a specific dataset (with fixed image dimensions) and
thus did not specify how to handle different image sizes and aspect ratios during inference. We tried to find the
best-performing setting for all methods, following their evaluation scripts and training dimensions.
For approaches trained on square patches~\cite{Xian2018}, we follow our setup and set the larger axis to the training image axis length and adapt the smaller one, keeping the aspect ratio as close as possible to the original. For approaches with non-square patches~\cite{Chen2016,LiSnavely2018,Li2019,Wang2019} we fix the smaller axis to the smaller training image axis dimension.
For DORN~\cite{Fu2018} we followed their tiling protocol, resizing the images to the dimensions stated for their NYU and KITTI evaluation, respectively.
For Monodepth2~\cite{Godard2019} and Struct2Depth~\cite{Casser2019}, which were both trained on KITTI and thus expect a very wide
aspect ratio, we pad the input image \changed{using reflection padding} to obtain the same aspect ratio, resize to their specific input dimension, and crop the resulting prediction to the original target dimensions.
For methods where model weights were available for different training resolutions we evaluated all of them and report numbers for the
best-performing variant.

All predictions were rescaled to the resolution of the ground truth for evaluation.

\begin{table}[b]
  \caption{Relative performance with respect to the baseline in percent when fine-tuning on different single
    training sets (higher is better). \changed{Performance better than the baseline in green, worse performance in red. Best performance is bold, second best is underlined.} The absolute errors of the RW baseline are shown on the top row.
    While some datasets provide better performance on individual, similar datasets, average
    performance for zero-shot cross-dataset transfer degrades.}
  \label{tab:singleDS}
    \robustify\bfseries
    \robustify\uline
    \newcolumntype{R}[0]{S[table-format=3.1,table-number-alignment = right,round-mode=places,round-precision=1,detect-weight,detect-mode]}
    \newcolumntype{Q}[0]{S[table-format=3.1,table-number-alignment = center,round-mode=places,round-precision=1,detect-mode]}
    \centering
  \scalebox{0.92}{
  \begin{tabular}{@{}l@{\hspace{1mm}}|@{\hspace{2mm}}RRRRRR@{\hspace{2mm}}|@{\hspace{2mm}}Q@{}}
\toprule
  & {DIW} & {ETH3D} &  {Sintel} & {KITTI} & {NYU} & {TUM} & {Mean [\%]}\\
\midrule
RW $\rightarrow$ RW   & \bfseries 14.59 & 0.151  & \bfseries 0.349 & \Uline{27.95} & \Uline{18.74} & 21.69 & \textemdash\\
\midrule
RW $\rightarrow$ DL  &   \red{-37.63} & \Uline{\green{1.99}}  & \red{-4.30} & \red{-72.99} &  \green{\bfseries 32.34} &  \green{\bfseries 19.41} &  \Uline{\red{-10.20}} \\
RW $\rightarrow$ MV   &   \Uline{\red{-26.05}} & \red{-15.89} & \red{-15.47} &  \green{\bfseries 10.13} & \red{-10.19} & \red{-3.46} & \Uline{\red{-10.16}} \\
RW $\rightarrow$ MD   &   \red{-31.46} & \green{\bfseries 3.97}  & \red{-9.74} & \red{-24.26} & \red{-1.65} & \red{-51.96} & \red{-19.18} \\
RW $\rightarrow$ WS   &   \red{-32.35} & \red{-29.80} & \Uline{\red{-2.87}} & \red{-34.49} & \red{-31.91} & \Uline{\green{3.23}} & \red{-21.37}\\
\bottomrule
\end{tabular}
}
\vspace*{\floatsep}
\caption{Absolute performance when fine-tuning on different single
    training sets  -- lower is better. This table corresponds to Table~\ref{tab:singleDS}.}
  \label{tab:singleDS_abs}
  \robustify\bfseries
    \robustify\uline
    \newcolumntype{R}[0]{S[table-format=2.2,table-number-alignment = center,round-mode=places,round-precision=2,detect-weight,detect-mode]}
    \newcolumntype{Q}[0]{S[table-format=2.3,table-number-alignment = center,round-mode=places,round-precision=3,detect-mode]}
    \centering
  \scalebox{0.95}{
  \begin{tabular}{@{}l@{\hspace{1mm}}|@{\hspace{2mm}}RQQRRR@{\hspace{2mm}}}
\toprule
  & {DIW} & {ETH3D} &  {Sintel} & {KITTI} & {NYU} & {TUM}\\
 & \changed{WHDR} &  \changed{AbsRel} & \changed{AbsRel} & \changed{$\delta\!\!>\!\!1.25$} & \changed{$\delta\!\!>\!\!1.25$} & \changed{$\delta\!\!>\!\!1.25$}  \\
\midrule
RW $\rightarrow$ RW   & \bfseries14.59 & 0.151  & \bfseries0.349 & \Uline{27.95} & \Uline{18.74} & 21.69\\
RW $\rightarrow$ DL  & 20.08 & \Uline{0.1478} & 0.3635 & 48.3534 & \bfseries12.6836 & \bfseries17.4843\\
RW $\rightarrow$ MV   & \Uline{18.39} & 0.1746 & 0.4032 & \bfseries25.1212 & 20.6531 & 22.4350\\
RW $\rightarrow$ MD   & 19.18 & \bfseries0.1450 & 0.3831 & 34.7292 & 19.0502 & 32.9595\\
RW $\rightarrow$ WS   & 19.31 & 0.1955 & \Uline{0.3586} & 37.5883 & 24.7176 & \Uline{20.9857}\\
\bottomrule
\end{tabular}
}
\end{table}

\mypara{Comparison of loss functions.}
We show the effect of different loss functions on the validation performance in Figure~\ref{fig:loss}.
We used RW to train networks with different losses.
For the ordinal loss (\cf\ Equation~\eqref{eq:ordloss2}), we sample 5,000 point pairs randomly~\cite{Xian2018}.
Where appropriate, we combine losses with the gradient regularization term \eqref{eq:lgrad}. We also test a scale-invariant, but not shift-invariant, MSE in disparity space $\lL_{simse}$ by fixing $t\!=\!0$ in~\eqref{eq:ssi}.
The model trained with $\lL_{ord}$ corresponds to our reimplementation of Xian \etal~\cite{Xian2018}.
Figure~\ref{fig:loss} shows that our proposed trimmed MAE loss yields the lowest validation error over all datasets.
We thus conduct all experiments that follow using $\lL_{ssitrim} + \lL_{reg}$.

\mypara{Comparison of encoders.}
We evaluate the influence of the encoder architecture in Figure~\ref{fig:encoder}. We define the model with a ResNet-50
\cite{He2016} encoder as used originally by Xian \etal~\cite{Xian2018} as our baseline and show the relative improvement in
performance when swapping in different encoders (higher is better). We tested ResNet-101, ResNeXt-101~\cite{Xie2017} and DenseNet-161
\cite{Huang2017}. All encoders were pretrained on ImageNet~\cite{Deng2009}. For ResNeXt-101, we additionally use a variant that was
pretrained with a massive corpus of weakly-supervised data (WSL)~\cite{Mahajan2018} before training on ImageNet. All models were fine-tuned on RW.

We observe that a significant performance boost is achieved by using better encoders.
Higher-capacity encoders perform better than the baseline.
The {ResNeXt-101} encoder that was pretrained on weakly-supervised data performs significantly better than the same encoder
that was only trained on ImageNet. \changed{We found pretraining to be crucial. A network with a ResNet-50 encoder with random initialization performs on average 35\% worse than its pretrained counterpart.}
In general, we find that ImageNet performance of an encoder is a strong predictor for its
performance in monocular depth estimation. This is encouraging, since advancements made in image classification
can directly yield gains in robust monocular depth estimation.
The performance gain over the baseline is remarkable: up to 15 \% relative improvement, without any task-specific adaptations.
We use {ResNeXt-101-WSL} for all subsequent experiments.

\begin{figure*}[!t]
  \centering
  \def\mywidth{0.16}
  \begin{tabular}{@{}c@{\hspace{1mm}}c@{\hspace{1mm}}c@{\hspace{1mm}}c@{\hspace{1mm}}c@{\hspace{1mm}}c@{}}
    Input & MIX 1 & MIX 2 & MIX 3 & MIX 4 & MIX 5\\
    \adjincludegraphics[width=\mywidth\linewidth,trim={0 0 0 0},clip]{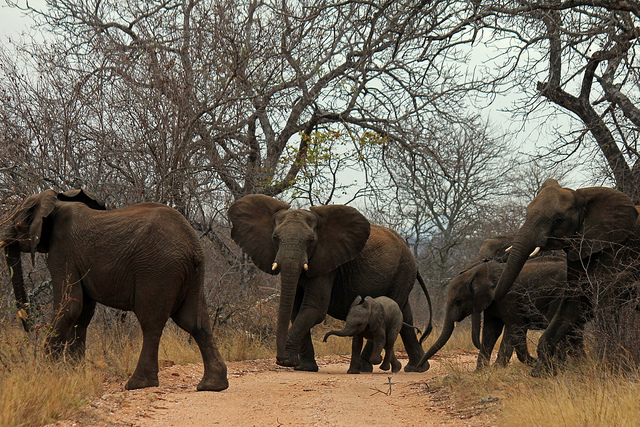} &
    \adjincludegraphics[width=\mywidth\linewidth,trim={0 0 0 0},clip]{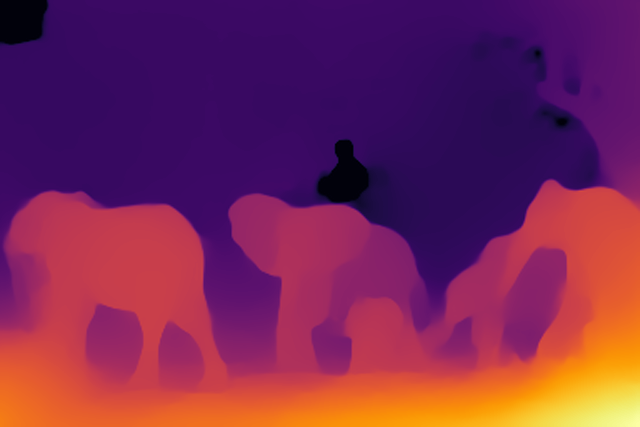} &
    \adjincludegraphics[width=\mywidth\linewidth,trim={0 0 0 0},clip]{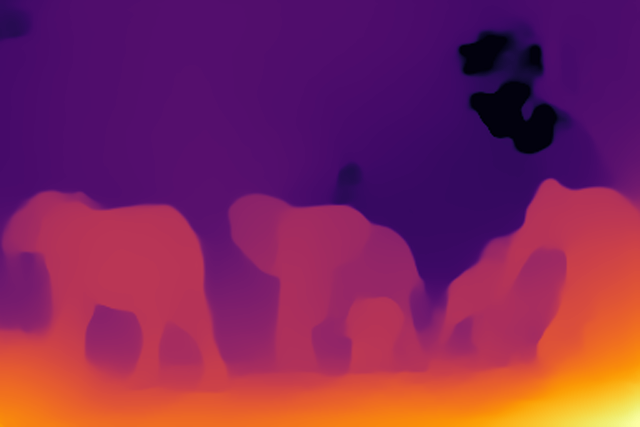} &
    \adjincludegraphics[width=\mywidth\linewidth,trim={0 0 0 0},clip]{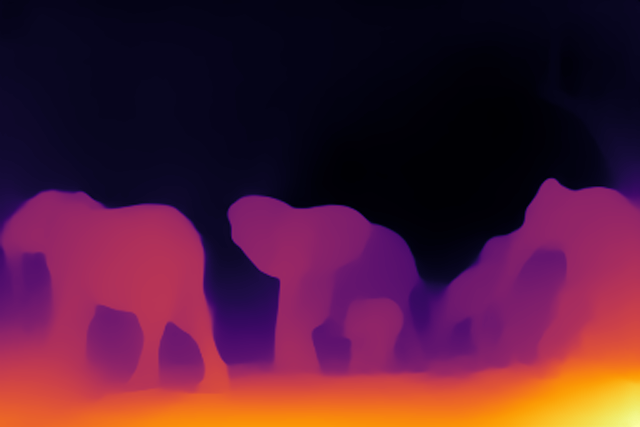} &
    \adjincludegraphics[width=\mywidth\linewidth,trim={0 0 0 0},clip]{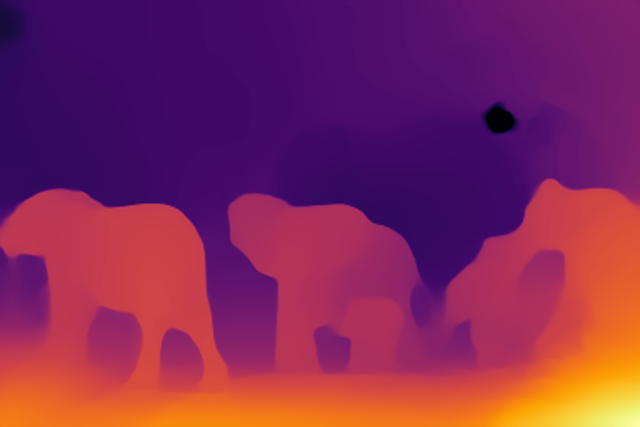} &
    \adjincludegraphics[width=\mywidth\linewidth,trim={0 0 0 0},clip]{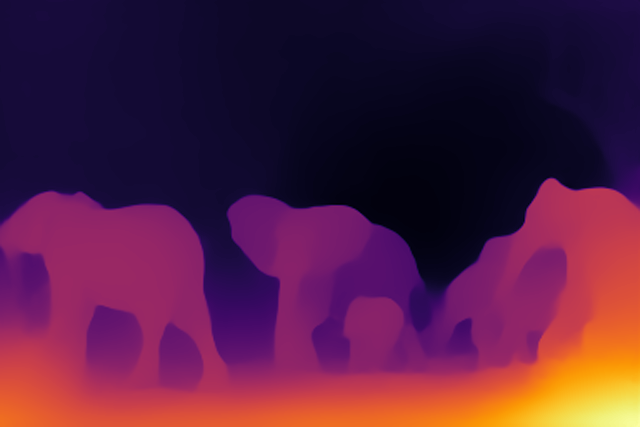} \\
    \adjincludegraphics[width=\mywidth\linewidth,trim={0 0 0 0},clip]{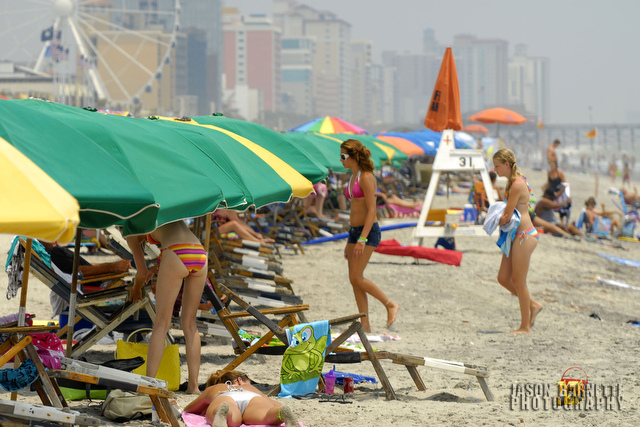} &
    \adjincludegraphics[width=\mywidth\linewidth,trim={0 0 0 0},clip]{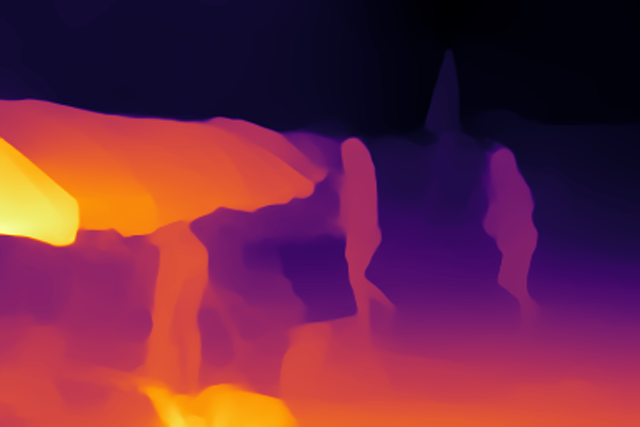} &
    \adjincludegraphics[width=\mywidth\linewidth,trim={0 0 0 0},clip]{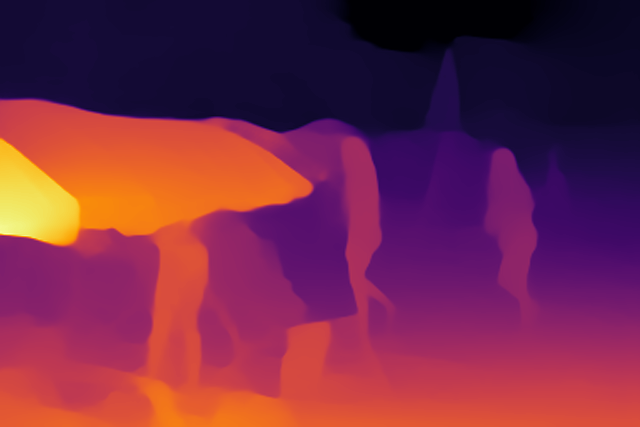} &
    \adjincludegraphics[width=\mywidth\linewidth,trim={0 0 0 0},clip]{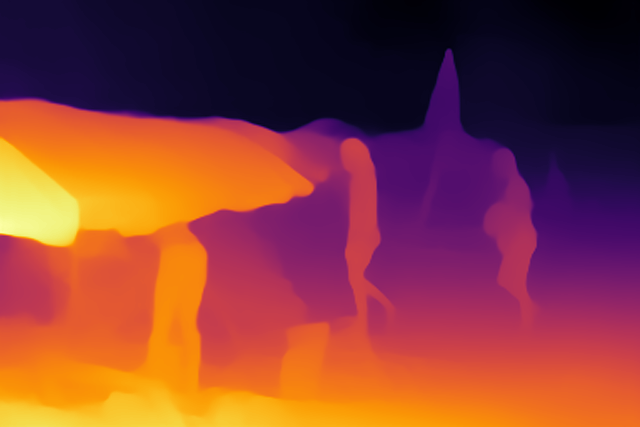} &
    \adjincludegraphics[width=\mywidth\linewidth,trim={0 0 0 0},clip]{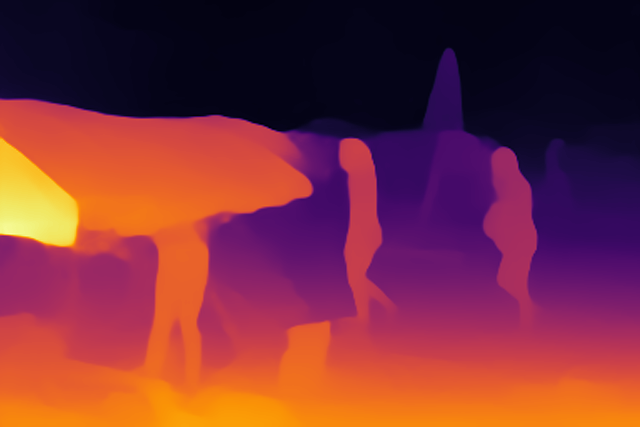} &
    \adjincludegraphics[width=\mywidth\linewidth,trim={0 0 0 0},clip]{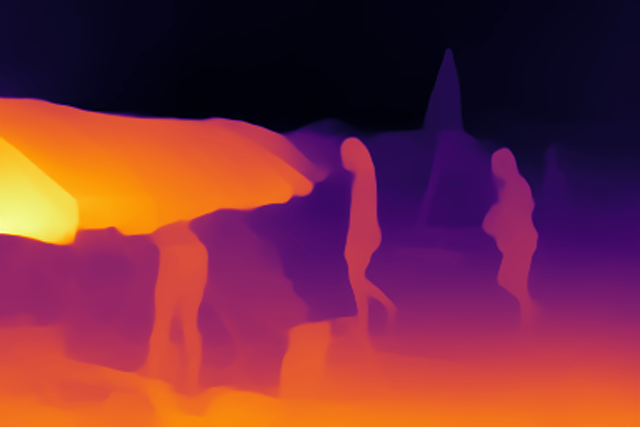} \\
    \adjincludegraphics[width=\mywidth\linewidth,trim={0 0 0 0},clip]{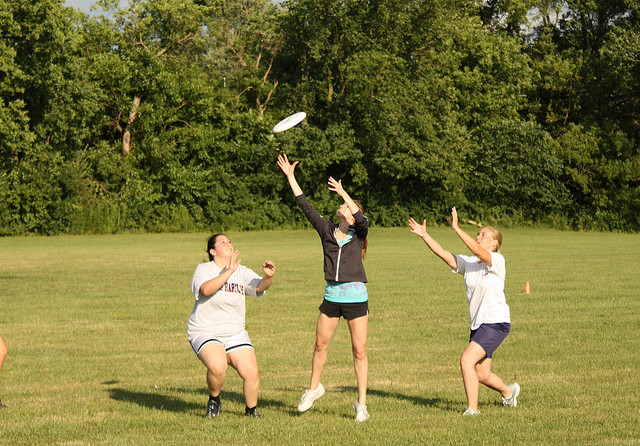} &
    \adjincludegraphics[width=\mywidth\linewidth,trim={0 0 0 0},clip]{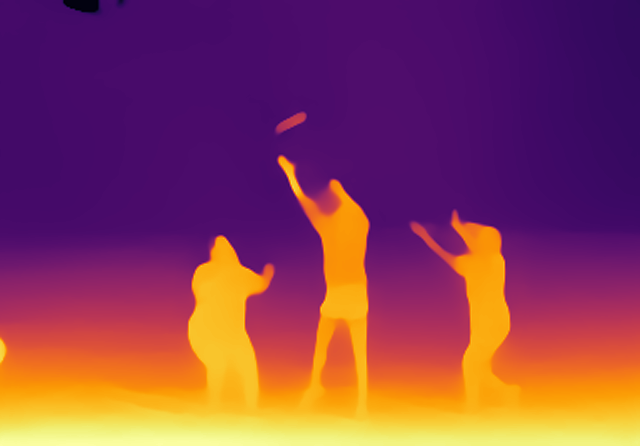} &
    \adjincludegraphics[width=\mywidth\linewidth,trim={0 0 0 0},clip]{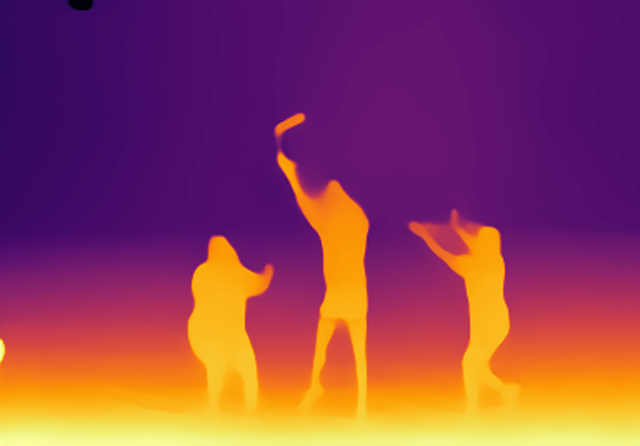} &
    \adjincludegraphics[width=\mywidth\linewidth,trim={0 0 0 0},clip]{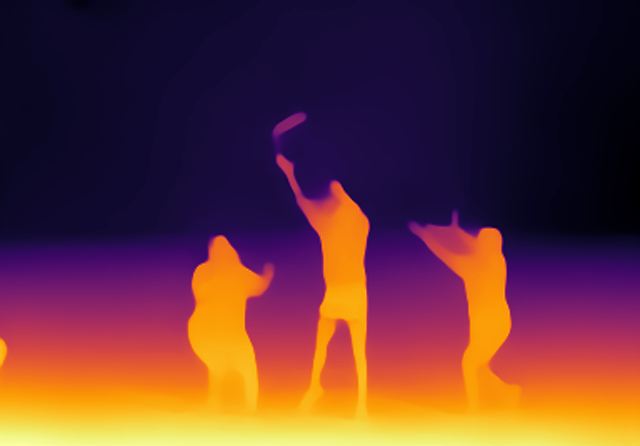} &
    \adjincludegraphics[width=\mywidth\linewidth,trim={0 0 0 0},clip]{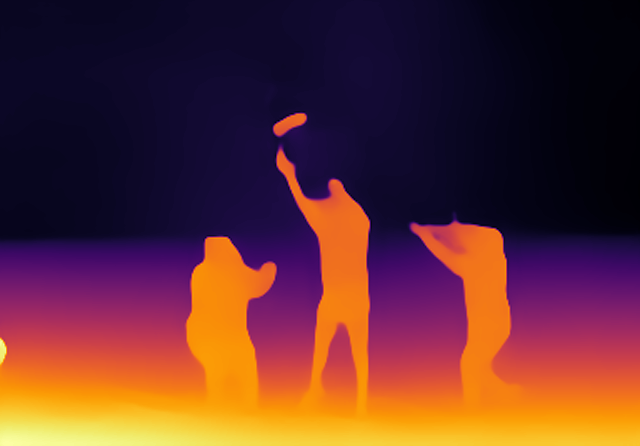} &
    \adjincludegraphics[width=\mywidth\linewidth,trim={0 0 0 0},clip]{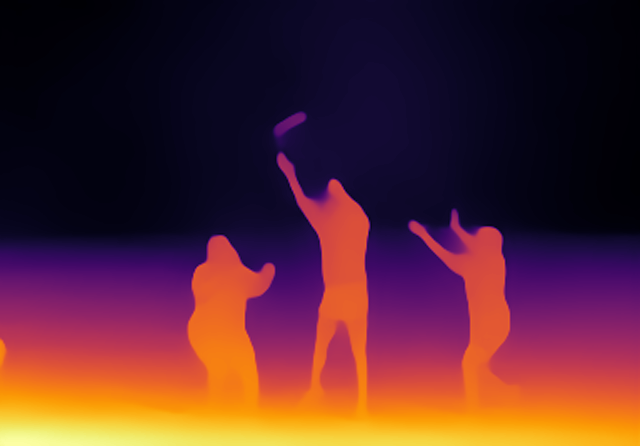} \\
\end{tabular}
\caption{\changed{Comparison of models trained on different combinations of datasets using Pareto-optimal mixing. Images from Microsoft COCO \cite{Lin2014}.}} %
\label{fig:results_comparison}
\end{figure*}

\begin{figure*}[!t]
  \centering
  \def\mywidth{0.19}
  \begin{tabular}{@{}c@{\hspace{1mm}}c@{\hspace{1mm}}c@{\hspace{1mm}}c@{\hspace{1mm}}c@{}}
    \raisebox{\height}{\rot{\small  Input}}
    \adjincludegraphics[width=\mywidth\linewidth,trim={0 0 0 0},clip]{figures/results/coco/selection/000000022935} &
    \adjincludegraphics[width=\mywidth\linewidth,trim={0 0 {0.11\width} 0},clip]{figures/results/coco/selection/000000405691} &
    \adjincludegraphics[width=\mywidth\linewidth,trim={{0.112\width} 0 0 0},clip]{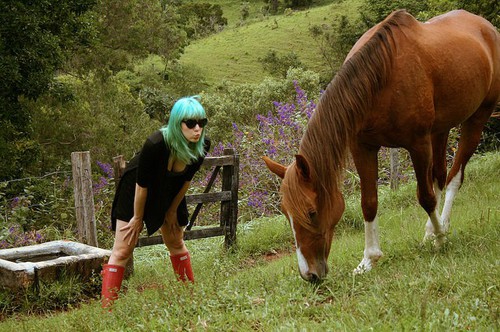} &
    \adjincludegraphics[width=\mywidth\linewidth,trim={0 0 {0.05\width} 0},clip]{figures/results/coco/selection/000000345361} &
    \adjincludegraphics[width=\mywidth\linewidth,trim={{0.07\width} 0 0 0},clip]{figures/results/coco/selection/000000030213}
    \\
    \raisebox{.15\height}{\rot{\small Xian \etal \cite{Xian2018}}}
    \adjincludegraphics[width=\mywidth\linewidth,trim={0 0 0 0},clip]{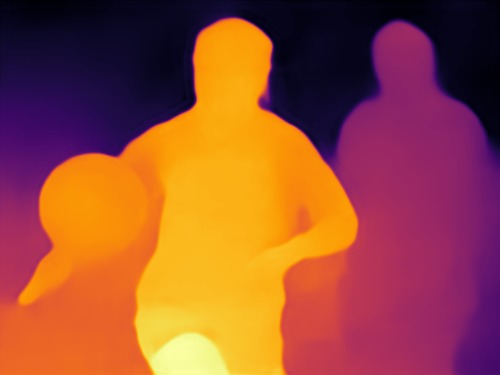} &
    \adjincludegraphics[width=\mywidth\linewidth,trim={0 0 {0.11\width} 0},clip]{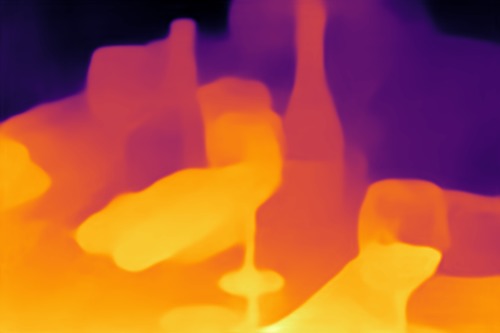} &
    \adjincludegraphics[width=\mywidth\linewidth,trim={{0.112\width} 0 0 0},clip]{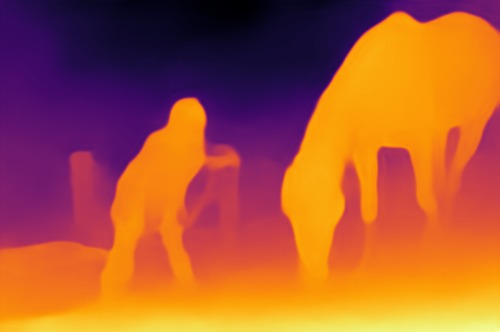} &
    \adjincludegraphics[width=\mywidth\linewidth,trim={0 0 {0.05\width} 0},clip]{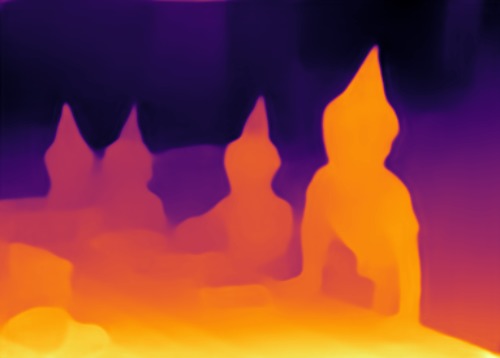} &
    \adjincludegraphics[width=\mywidth\linewidth,trim={{0.07\width} 0 0 0},clip]{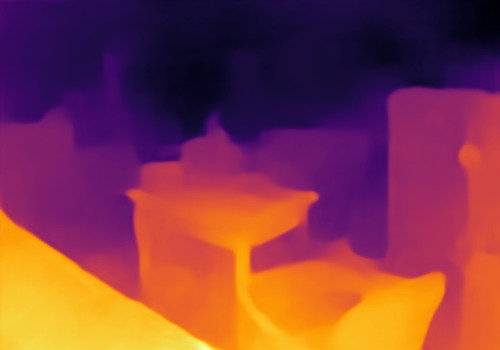}
    \\
     \raisebox{.15\height}{\rot{\small  Wang \etal \cite{Wang2019}}}
    \adjincludegraphics[width=\mywidth\linewidth,trim={0 0 0 0},clip]{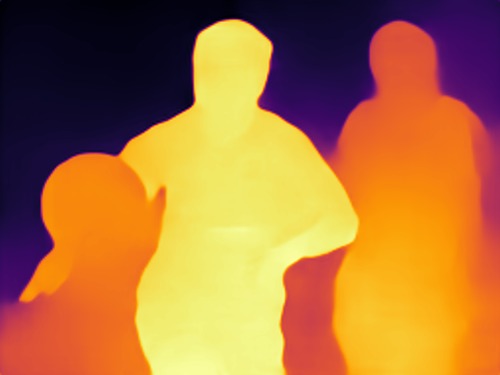} &
    \adjincludegraphics[width=\mywidth\linewidth,trim={0 0 {0.11\width} 0},clip]{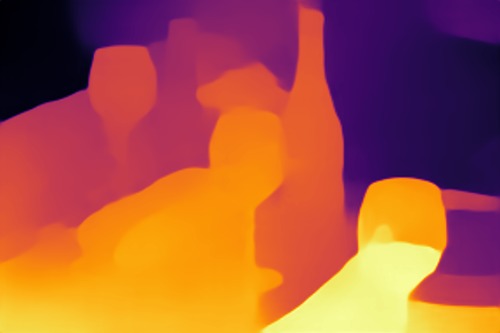} &
    \adjincludegraphics[width=\mywidth\linewidth,trim={{0.112\width} 0 0 0},clip]{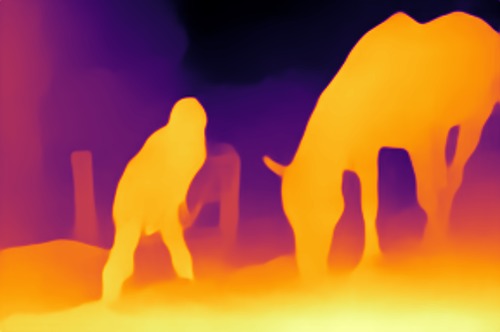} &
    \adjincludegraphics[width=\mywidth\linewidth,trim={0 0 {0.05\width} 0},clip]{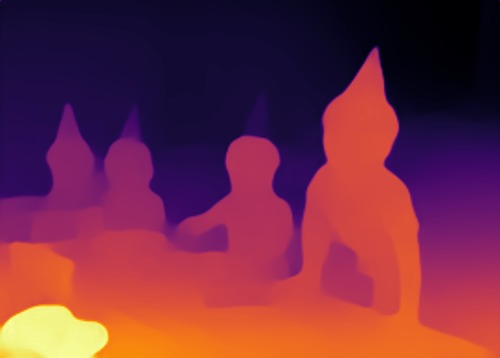} &
    \adjincludegraphics[width=\mywidth\linewidth,trim={{0.07\width} 0 0 0},clip]{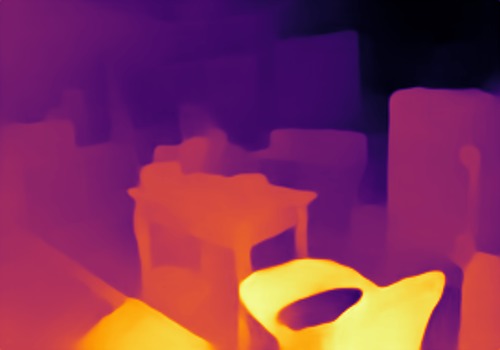}
    \\
    \raisebox{.27\height}{\rot{\small  Li \etal \cite{Li2019}}}
    \adjincludegraphics[width=\mywidth\linewidth,trim={0 0 0 0},clip]{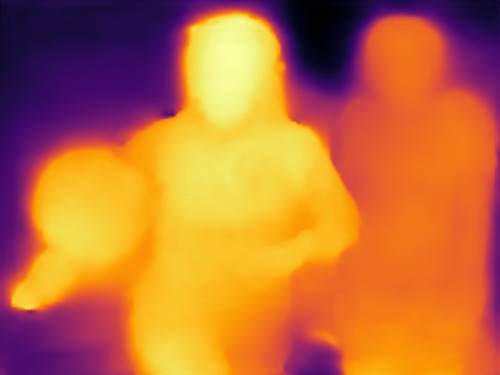} &
    \adjincludegraphics[width=\mywidth\linewidth,trim={0 0 {0.11\width} 0},clip]{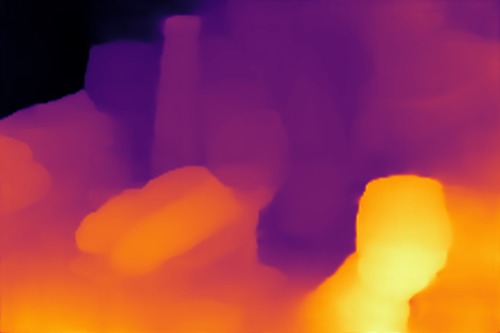} &
    \adjincludegraphics[width=\mywidth\linewidth,trim={{0.112\width} 0 0 0},clip]{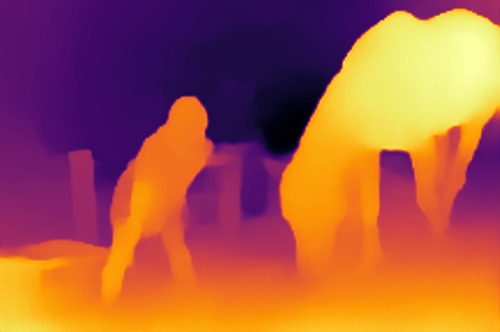} &
    \adjincludegraphics[width=\mywidth\linewidth,trim={0 0 {0.05\width} 0},clip]{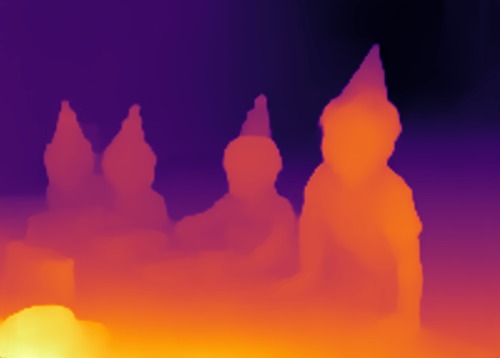} &
    \adjincludegraphics[width=\mywidth\linewidth,trim={{0.07\width} 0 0 0},clip]{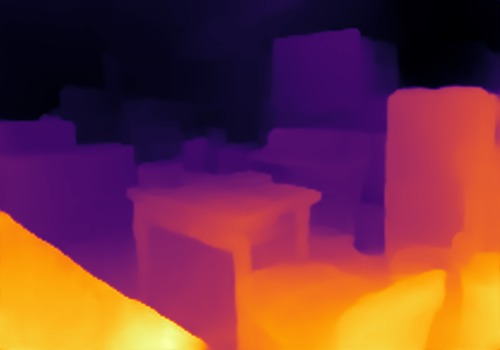}
    \\
    \raisebox{0.04\height}{\rot{\small  Li \& Snavely \cite{LiSnavely2018}}}
    \adjincludegraphics[width=\mywidth\linewidth,trim={0 0 0 0},clip]{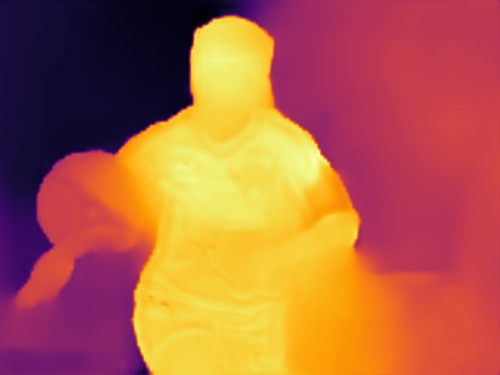} &
    \adjincludegraphics[width=\mywidth\linewidth,trim={0 0 {0.11\width} 0},clip]{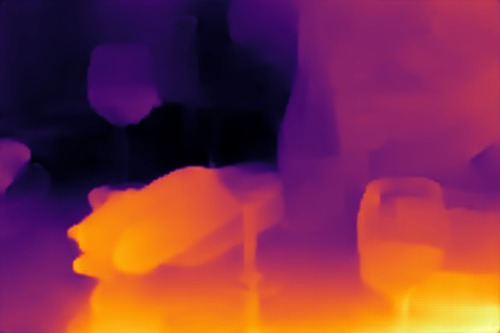} &
    \adjincludegraphics[width=\mywidth\linewidth,trim={{0.112\width} 0 0 0},clip]{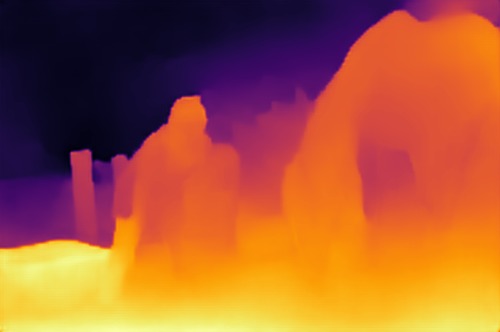} &
    \adjincludegraphics[width=\mywidth\linewidth,trim={0 0 {0.05\width} 0},clip]{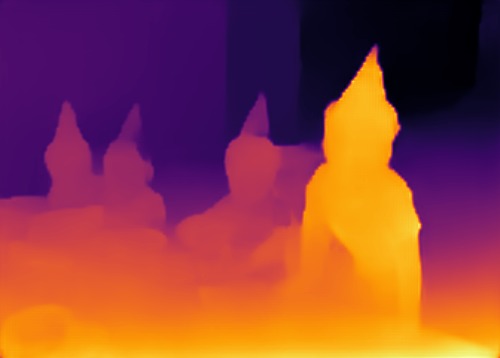} &
    \adjincludegraphics[width=\mywidth\linewidth,trim={{0.07\width} 0 0 0},clip]{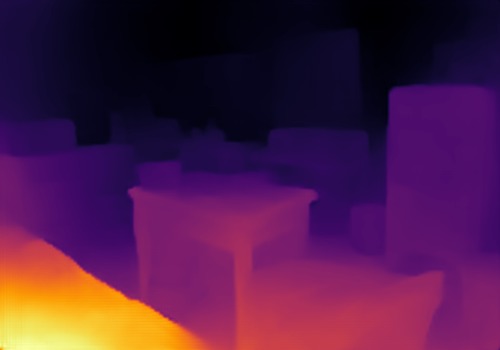}
    \\
    \raisebox{1.5\height}{\rot{\small Ours}}
    \adjincludegraphics[width=\mywidth\linewidth,trim={0 0 0 0},clip]{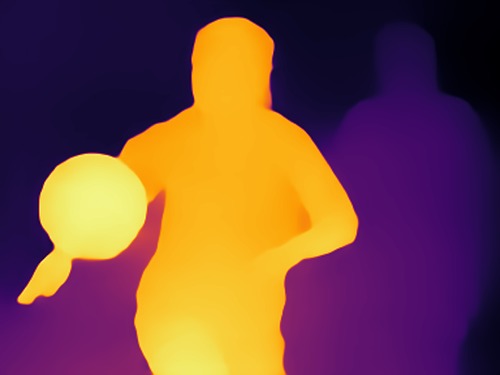} &
    \adjincludegraphics[width=\mywidth\linewidth,trim={0 0 {0.11\width} 0},clip]{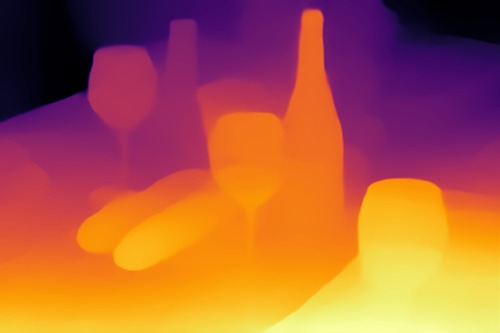} &
    \adjincludegraphics[width=\mywidth\linewidth,trim={{0.112\width} 0 0 0},clip]{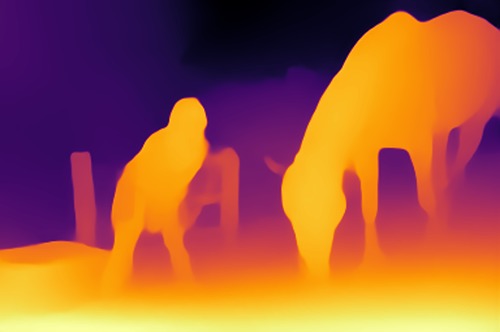} &
    \adjincludegraphics[width=\mywidth\linewidth,trim={0 0 {0.05\width} 0},clip]{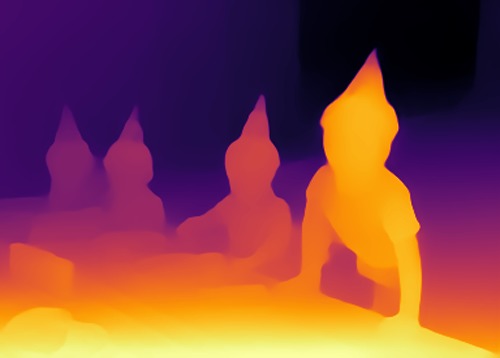} &
    \adjincludegraphics[width=\mywidth\linewidth,trim={{0.07\width} 0 0 0},clip]{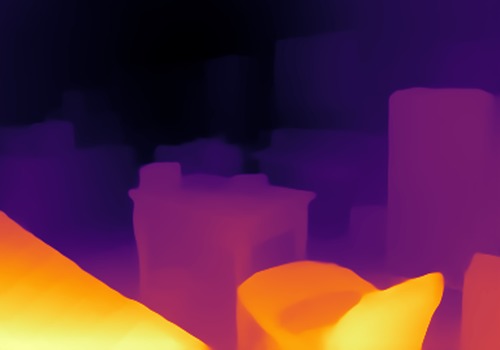}

\end{tabular}
\caption{Qualitative comparison of our approach to the four best competitors on images from the Microsoft COCO dataset \cite{Lin2014}.} %
\label{fig:results_movies}
\end{figure*}

\begin{table}[t]
  \caption{Combinations of datasets used for training.}%
  \label{tab:dataset_mix}
  \robustify\bfseries
    \newcolumntype{R}[0]{S[table-format=3.1,table-number-alignment = right,round-mode=places,round-precision=1,detect-weight,detect-mode]}
    \newcolumntype{Q}[0]{S[table-format=3.1,table-number-alignment = center,round-mode=places,round-precision=1,detect-mode]}
  \centering
  \scalebox{0.95}{
  \ra{1.05}
\begin{tabular}{@{}cc@{\hspace{2mm}}c@{\hspace{2mm}}c@{\hspace{2mm}}c@{\hspace{2mm}}c@{\hspace{2mm}}c@{\hspace{2mm}}}%

  \toprule
Mix & RW    & DL     & MV    & MD    & WS    \\ %

\midrule
  MIX 1 &\YesV & \YesV &         &          &          \\%&     \\
  MIX 2 &\YesV & \YesV &   \YesV   &          &       \\  %
  MIX 3 &\YesV & \YesV & \YesV & \YesV &   &    \\
  \changed{MIX 4} & \changed{\YesV} & \changed{\YesV} &  &\changed{\YesV} & \changed{\YesV} \\%  \\
  \changed{MIX 5} & \YesV & \YesV & \YesV &\YesV & \YesV \\%  \\
\bottomrule
\end{tabular}}
\end{table}

\begin{table}[b!]
  \caption{Relative performance of naive dataset mixing with respect to the RW baseline (top row) -- higher is
    better.
    While we usually see an improvement when adding datasets, adding datasets can hurt generalization performance
    with naive mixing.}
  \label{tab:mixing}
  \robustify\bfseries
    \robustify\uline
    \newcolumntype{R}[0]{S[table-format=3.1,table-number-alignment = center,round-mode=places,round-precision=1,detect-weight,detect-mode]}
    \newcolumntype{Q}[0]{S[table-format=3.1,table-number-alignment = center,round-mode=places,round-precision=1,detect-mode]}
  \centering
  \scalebox{0.95}{
  \ra{1.05}
\begin{tabular}{@{}l@{\hspace{2mm}}|RRRRRR@{\hspace{2mm}}|Q@{}}

  \toprule
  &                    {DIW} &                {ETH3D} &             {Sintel} &
{KITTI} &               {NYU} &            {TUM}& {Mean [\%]}\\
\midrule
RW   &  14.59 & 0.151  & 0.349 & 27.95 & 18.74 & 21.69 & \textemdash\\
\midrule
MIX 1   &        \green{10.90} &  \green{9.93} &  \red{-3.72} & \green{\bfseries18.03} &  \Uline{\green{41.41}} & \green{33.01} & \green{18.26}\\
MIX 2   &      \green{6.65} &  \green{8.61} &   \green{3.15} &  \green{9.16} & \green{40.77} &  \Uline{\green{35.73}} & \green{17.34}\\
MIX 3  &       \green{\bfseries13.50} &  \green{10.60} &    \green{4.87} &  \Uline{\green{13.92}} & \green{\bfseries43.76} & \green{29.05} &  \Uline{\green{19.28}}\\
\changed{MIX 4} & \changedg{11.7} & \Uline{\changedg{11.3}} & \Uline{\changedg{5.2}} & \changedg{11.3} & \changedg{38.8} & \changedg{35.5} & \changedg{19.0} \\
\changed{MIX 5} &        \Uline{\green{12.34}} & \green{\bfseries12.58}  &  \green{\bfseries7.16}  & \green{9.09} & \green{38.53} & \green{\bfseries37.21} & \green{\bfseries19.48}\\
\bottomrule
\end{tabular}}
    \vspace*{\floatsep}
\caption{Absolute performance of naive dataset mixing -- lower is
    better. This table corresponds to Table~\ref{tab:mixing}.}
  \label{tab:mixing_abs}
  \robustify\bfseries
    \newcolumntype{R}[0]{S[table-format=2.2,table-number-alignment = center,round-mode=places,round-precision=2,detect-weight,detect-mode]}
    \newcolumntype{Q}[0]{S[table-format=2.3,table-number-alignment = center,round-mode=places,round-precision=3,detect-mode]}
    \centering
  \scalebox{0.95}{
  \ra{1.05}
\begin{tabular}{@{}l@{\hspace{2mm}}|RQQRRR@{\hspace{2mm}}}

  \toprule
  &                    {DIW} &                {ETH3D} &             {Sintel} &
{KITTI} &               {NYU} &            {TUM} \\
 & \changed{WHDR} &  \changed{AbsRel} & \changed{AbsRel} & \changed{$\delta\!\!>\!\!1.25$} & \changed{$\delta\!\!>\!\!1.25$} & \changed{$\delta\!\!>\!\!1.25$}  \\
\midrule
RW   &  14.59 & 0.151  & 0.349 & 27.95 & 18.74 & 21.69\\
MIX 1 & 13.00 & 0.1362 & 0.3622 & \bfseries22.9094 & \Uline{10.9767} & 14.5313\\
MIX 2 & 13.62 & 0.1379 & 0.3378 & 25.3891 & 11.0968 & \Uline{13.9384}\\
MIX 3 & \bfseries12.62 & 0.1347 & 0.3323 & \Uline{24.0644} & \bfseries10.5406 & 15.3931\\
\changed{MIX 4}  & \changedt{12.88} & \Uline{\changedt{0.134}} & \Uline{\changedt{0.331}} & \changedt{24.78} & \changedt{11.46} & \changedt{14.00} \\
  \changed{MIX 5} & \Uline{12.79} & \bfseries0.1324 & \bfseries0.3243 & 25.4147 & 11.5185 & \bfseries13.6244\\
\bottomrule
  \end{tabular}}
\end{table}

\mypara{Training on diverse datasets.}
We evaluate the usefulness of different training datasets for generalization in Table~\ref{tab:singleDS} and Table~\ref{tab:singleDS_abs}. While more
specialized datasets reach better performance on similar test sets (DL for indoor scenes or MD for ETH3D), performance on the remaining datasets declines. Interestingly, every single dataset used in isolation leads to
worse generalization performance on average than just using the small, but curated, RW dataset,~\ie\ the gains on
compatible datasets are offset on average by the decrease on the other datasets.

The difference in performance for RW, MV, and WS is especially interesting since they have similar characteristics.
Although substantially larger than RW, both MV and WS show worse individual performance. This could be explained partly by
redundant data due to the video nature of these datasets and possibly more rigorous filtering in RW (human
experts pruned samples that had obvious flaws).
Comparing WS and MV, we see that MV leads to more general models, likely because of higher-quality stereo pairs due to
the more controlled nature of the images.

For our subsequent mixing experiments, we use Table \ref{tab:singleDS} as reference, \ie\ we start with the best performing
individual training dataset and consecutively add datasets to the mix. We show which datasets are included in the
individual training sets in Table \ref{tab:dataset_mix}. \changed{To better understand the influence of the Movies dataset, we additionally show results where we train on all datasets except Movies (MIX 4).} We always start training from the pretrained RW baseline.

Tables~\ref{tab:mixing} and \ref{tab:mixing_abs} show that, in contrast to using individual datasets, mixing multiple training sets consistently
improves performance with respect to the baseline. However, we also see that adding datasets does not unconditionally
improve performance when naive mixing is used (see MIX 1 vs.\ MIX 2).
Tables \ref{tab:mgda} and \ref{tab:mgda_abs} report the results of an analogous experiment with Pareto-optimal dataset mixing. We
observe that this approach improves over the naive mixing strategy. It is also more consistently able to leverage
additional datasets. Combining all five datasets with Pareto-optimal mixing yields our best-performing model. \changed{We show a qualitative comparison
of the resulting models in Figure \ref{fig:results_comparison}.}

\begin{table}[t!]
  \caption{Relative performance of dataset mixing with multi-objective optimization with respect to the RW baseline
    (top row) -- higher is  better.
    Principled mixing dominates the solutions found by naive mixing.}
  \label{tab:mgda}
  \robustify\bfseries
    \robustify\uline
    \newcolumntype{R}[0]{S[table-format=3.1,table-number-alignment = center,round-mode=places,round-precision=1,detect-weight,detect-mode]}
    \newcolumntype{Q}[0]{S[table-format=3.1,table-number-alignment = center,round-mode=places,round-precision=1,detect-mode]}
  \centering
  \scalebox{0.95}{
  \ra{1.05}

\begin{tabular}{@{}l@{\hspace{2mm}}|RRRRRR@{\hspace{2mm}}|Q@{}}
\toprule
&                     {DIW} &                {ETH3D} &             {Sintel} &         {KITTI} &
{NYU} & {TUM} &  {Mean [\%]} \\
\midrule
RW    &  14.59 & 0.151  & 0.349 & 27.95 & 18.74 & 21.69 & \textemdash\\
\midrule
MIX 1   &   \green{9.39} &  \green{7.28} &  \red{-7.74} & \green{13.20} & \green{44.08} & \green{33.15} & \green{16.56}\\
MIX 2   &   \green{14.05} &  \green{8.61} &  \green{0.86} &  \green{\bfseries17.53} & \green{45.52} & \green{31.95} & \green{19.75}\\
MIX 3   &   \Uline{\green{15.76}} & \green{11.92} &  \Uline{\green{5.16}} & \green{11.70} &  \Uline{\green{47.81}} &  \green{32.41} &  \green{20.79}\\
\changed{MIX 4} & \changedg{15.4}  & \Uline{\changedg{13.9}} & \changedg{1.7} & \Uline{\changedg{17.2}} & \changedg{43.4} & \changedg{\bfseries38.2} & \Uline{\changedg{21.6}}\\
\changed{MIX 5}   &   \green{\bfseries15.90} & \green{\bfseries14.57}  &  \green{\bfseries6.30} & \green{14.49} & \green{\bfseries49.04} & \Uline{\green{34.12}} & \green{\bfseries22.40}\\
\bottomrule
\end{tabular}}

\vspace*{\floatsep}
  \caption{Absolute performance of dataset mixing with multi-objective optimization -- lower is
    better. This table corresponds to Table~\ref{tab:mgda}.}
  \label{tab:mgda_abs}

  \robustify\bfseries
    \newcolumntype{R}[0]{S[table-format=2.2,table-number-alignment = center,round-mode=places,round-precision=2,detect-weight,detect-mode]}
    \newcolumntype{Q}[0]{S[table-format=2.3,table-number-alignment = center,round-mode=places,round-precision=3,detect-mode]}

    \centering
  \scalebox{0.95}{
  \ra{1.05}
\begin{tabular}{@{}l@{\hspace{2mm}}|RQQRRR@{\hspace{2mm}}}

  \toprule
  &                    {DIW} &                {ETH3D} &             {Sintel} &
{KITTI} &               {NYU} &            {TUM} \\
 & \changed{WHDR} &  \changed{AbsRel} & \changed{AbsRel} & \changed{$\delta\!\!>\!\!1.25$} & \changed{$\delta\!\!>\!\!1.25$} & \changed{$\delta\!\!>\!\!1.25$}  \\
\midrule
RW   &  14.59 & 0.151 & 0.349 & 27.95 & 18.74 & 21.69\\
MIX 1 &  13.22 & 0.1401 & 0.3757 & 24.2593 & 10.4774 &  14.4979\\
MIX 2 &  12.54 & 0.1384 & 0.3462 & \bfseries23.0525 & 10.2116 & 14.7551\\
  MIX 3 &   \Uline{12.29} &  0.1334 &  \Uline{0.3310} & 24.6830 &   \Uline{9.7797} & 14.6606\\
\changed{MIX 4} & \changedt{12.35} & \Uline{\changedt{0.130}} & \changedt{0.343} & \Uline{\changedt{23.13}} & \changedt{10.61} & \changedt{\bfseries13.41} \\
  \changed{MIX 5} &  \bfseries12.27 & \bfseries0.1288 & \bfseries0.3273 &  23.8959 &  \bfseries9.5480 & \Uline{14.2920}\\
\bottomrule
  \end{tabular}}

\end{table}

\mypara{Comparison to the state of the art.}
We compare our best-performing model to various state-of-the-art approaches in Table~\ref{tab:sota} and Table~\ref{tab:baselines}. The top part of
each table compares to baselines that were not fine-tuned on any of the evaluated datasets (\ie\ zero-shot transfer, akin
to our model). The bottom parts show baselines that were fine-tuned on a subset of the datasets for reference.
In the training set column, MC refers to Mannequin Challenge~\cite{Li2019} and CS to Cityscapes~\cite{Cordts2016}.
\mbox{A $\rightarrow$ B} indicates pretraining on A and fine-tuning on B.

Our model outperforms the baselines by a comfortable margin in terms of zero-shot performance. Note that our model outperforms the Mannequin Challenge model of Li \etal~\cite{Li2019} on a subset of the TUM dataset that was specifically curated by Li \etal\ to showcase the advantages of their model.
\changed{We show additional results on a variant of our model that has a smaller encoder based on ResNet-50 (Ours -- small). This architecture
  is equivalent to the network proposed by Xian \etal~\cite{Xian2018}. The smaller model also outperforms the state of the art by a comfortable margin. This shows that the strong performance of our model is not only due to increased network capacity, but fundamentally due to the proposed training scheme.}

Some models that were trained for one specific dataset (\eg\ KITTI or NYU in the lower part of the table) perform very
well on those individual datasets but perform significantly worse on all other test sets. Fine-tuning on individual datasets
leads to strong priors about specific environments. This can be desirable in some applications, but is ill-suited if the
model needs to generalize.
A qualitative comparison of our model to the four best-performing competitors is shown in Figure~\ref{fig:results_movies}.

\begin{figure*}[htpb]
  \setlength{\linewidth}{\textwidth}
  \setlength{\hsize}{\textwidth}
  \def\mycolspace{1.2mm}
  \def\teaserwidth{0.245\linewidth}

  \centering
 	\begin{tabular}{@{}c@{\hspace{\mycolspace}}c@{\hspace{\mycolspace}}c@{\hspace{\mycolspace}}c@{}}
    \includegraphics[trim={0 0 0 0}, clip, width=\teaserwidth]{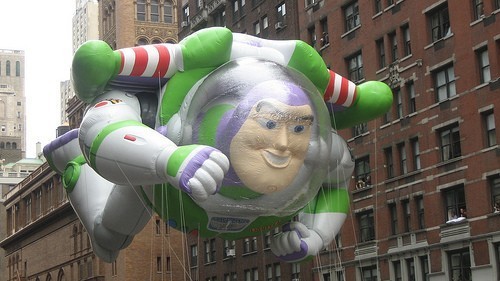} &
    \includegraphics[trim={0 0 0 0}, clip, width=\teaserwidth]{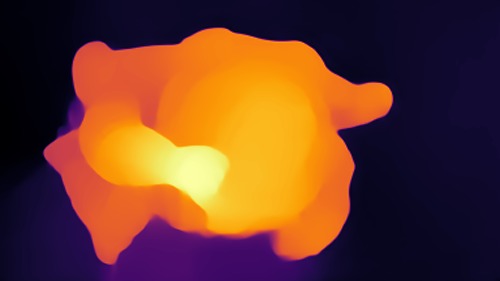} &
    \includegraphics[trim={0 0 0 0}, clip, width=\teaserwidth]{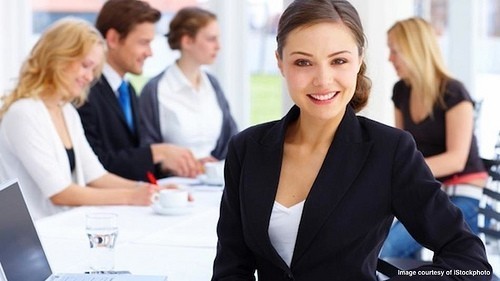} &
    \includegraphics[trim={0 0 0 0}, clip, width=\teaserwidth]{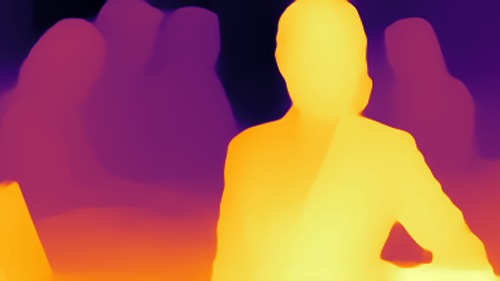} \\
        \includegraphics[trim={0 0 0 0}, clip, width=\teaserwidth]{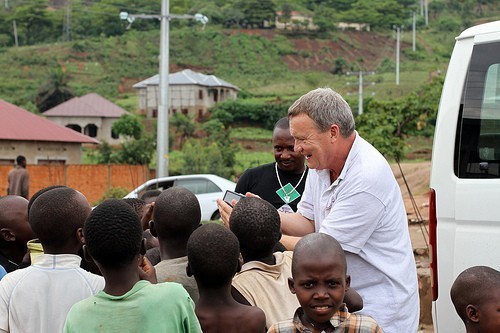} &
    \includegraphics[trim={0 0 0 0}, clip, width=\teaserwidth]{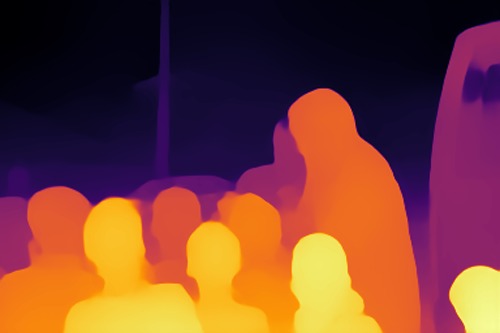} &
    \includegraphics[trim={0 0 0 0}, clip, width=\teaserwidth]{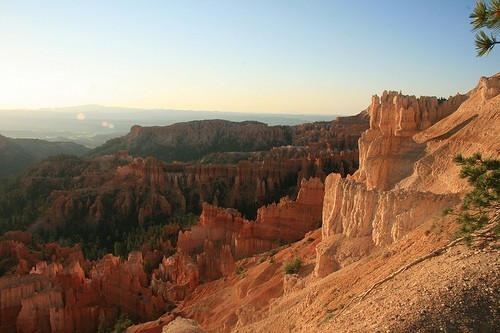} &
    \includegraphics[trim={0 0 0 0}, clip, width=\teaserwidth]{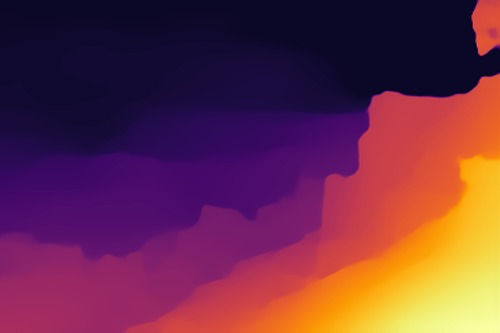} \\
                    \includegraphics[trim={0 0 0 0}, clip, width=\teaserwidth]{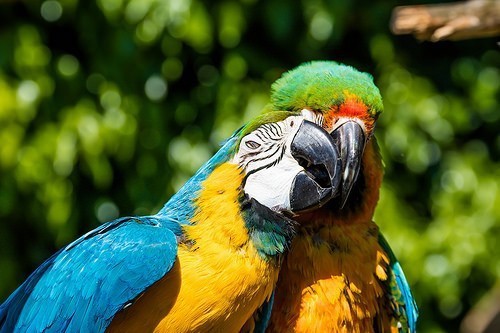} &
    \includegraphics[trim={0 0 0 0}, clip, width=\teaserwidth]{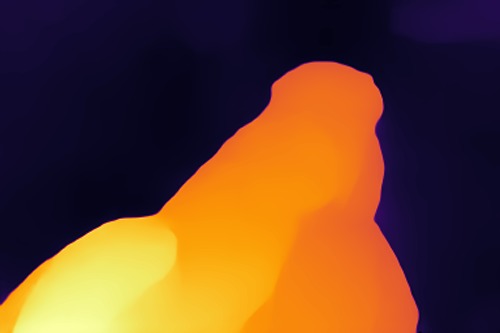} &
     \includegraphics[trim={0 0 0 0}, clip, width=\teaserwidth]{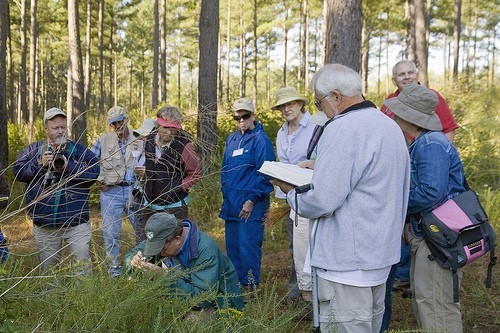} &
    \includegraphics[trim={0 0 0 0}, clip, width=\teaserwidth]{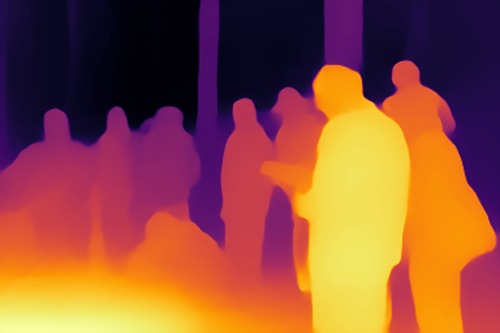} \\
        \includegraphics[trim={0 0 0 0}, clip, width=\teaserwidth]{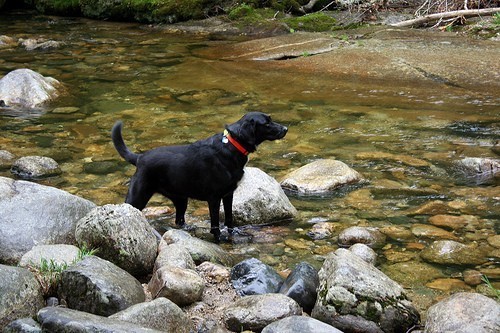} &
    \includegraphics[trim={0 0 0 0}, clip, width=\teaserwidth]{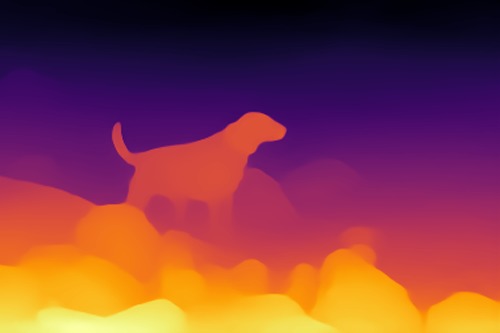} &
    \includegraphics[trim={0 0 0 0}, clip, width=\teaserwidth]{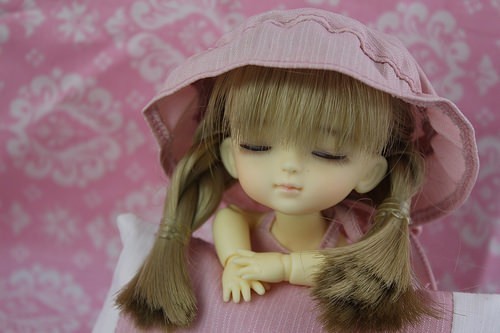} &
    \includegraphics[trim={0 0 0 0}, clip, width=\teaserwidth]{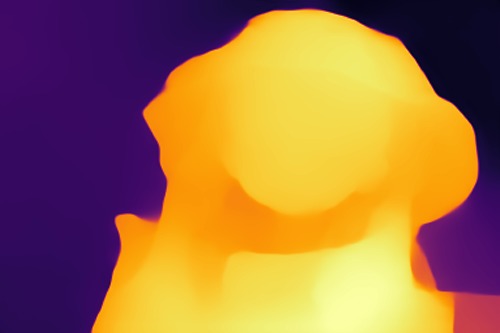} \\
          \includegraphics[trim={0 0 0 0}, clip, width=\teaserwidth, height=3.1cm]{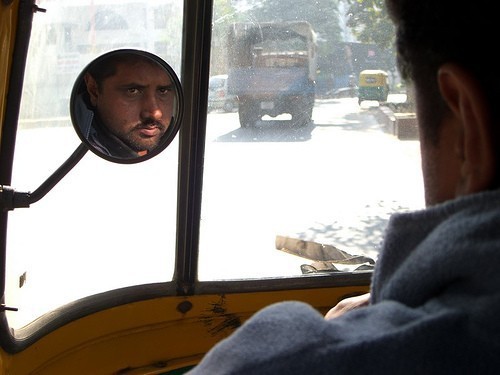} &
    \includegraphics[trim={0 0 0 0}, clip, width=\teaserwidth, height=3.1cm]{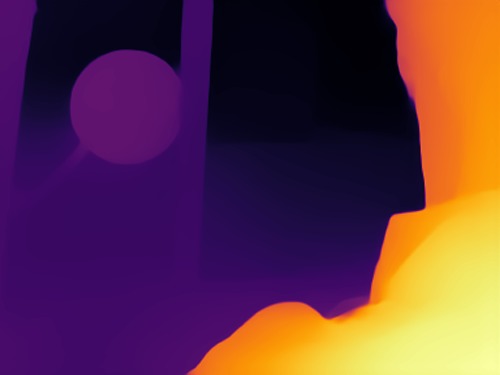} &
    \includegraphics[trim={0 0 0 0}, clip, width=\teaserwidth, height=3.1cm]{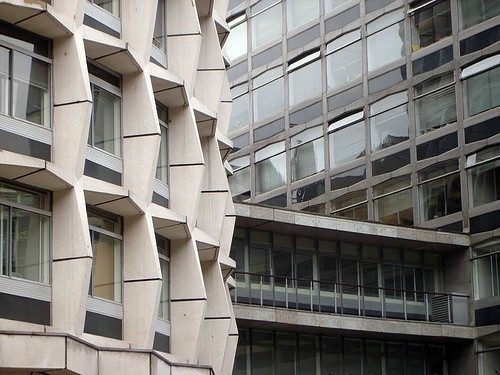} &
    \includegraphics[trim={0 0 0 0}, clip, width=\teaserwidth, height=3.1cm]{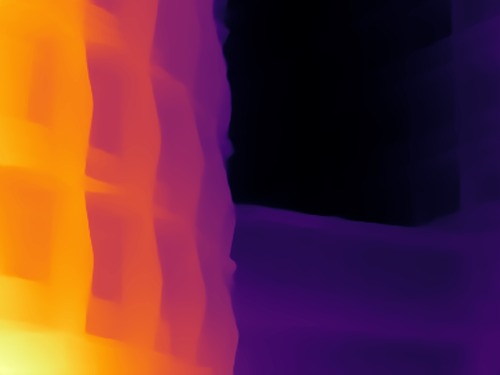} \\
            \includegraphics[trim={0 0 0 0}, clip, width=\teaserwidth, height=3.1cm]{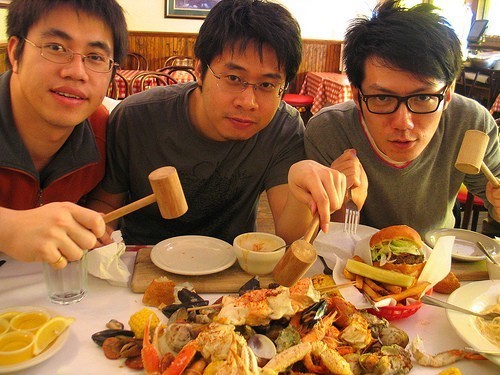} &
    \includegraphics[trim={0 0 0 0}, clip, width=\teaserwidth, height=3.1cm]{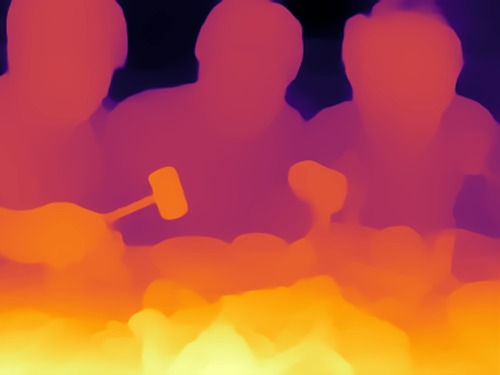} &
        \includegraphics[trim={0 0 0 0}, clip, width=\teaserwidth, height=3.1cm]{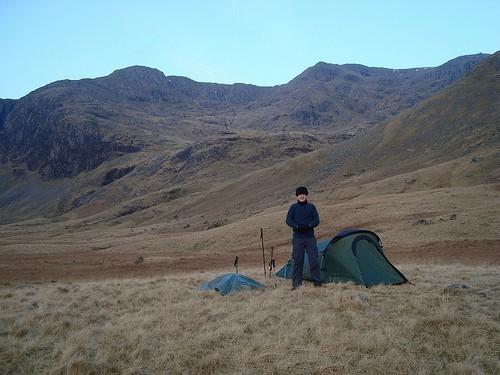} &
    \includegraphics[trim={0 0 0 0}, clip, width=\teaserwidth,                                                                                                                                               height=3.1cm]{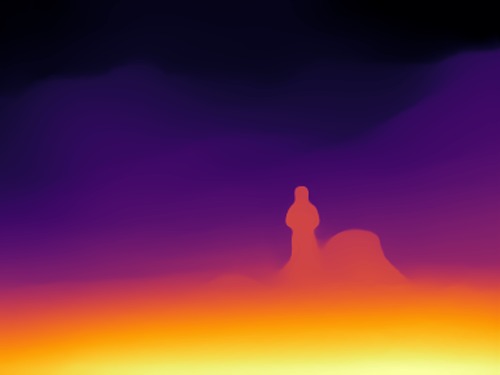} \\
      \includegraphics[trim={0 0 0 0}, clip, width=\teaserwidth, height=3.1cm]{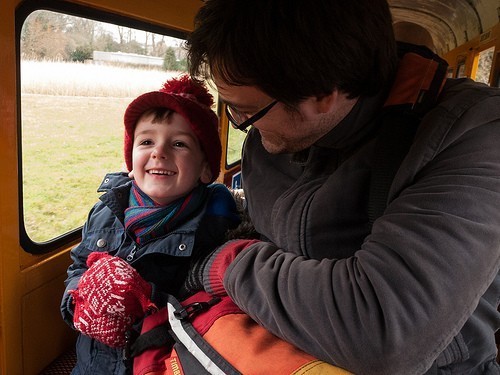} &
    \includegraphics[trim={0 0 0 0}, clip, width=\teaserwidth, height=3.1cm]{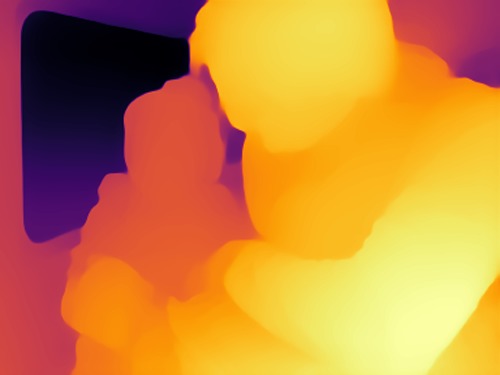} &
     \includegraphics[trim={0 0 0 0}, clip, width=\teaserwidth, height=3.1cm]{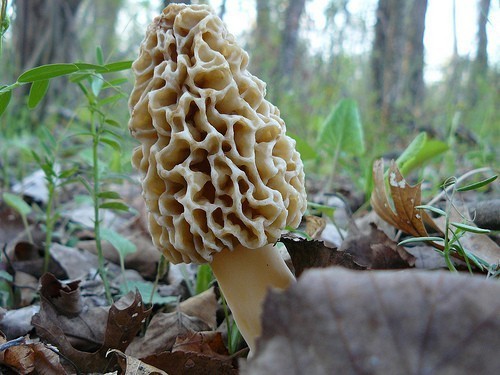} &
    \includegraphics[trim={0 0 0 0}, clip, width=\teaserwidth, height=3.1cm]{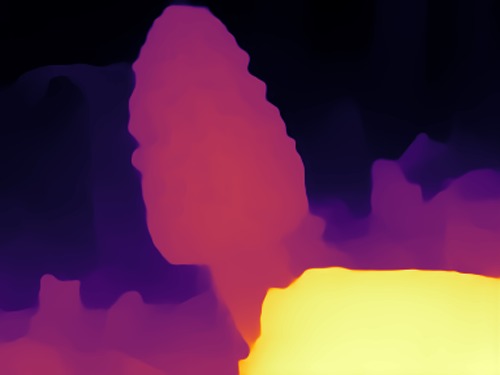} \\
              \includegraphics[trim={0 0 0 0}, clip, width=\teaserwidth, height=3cm]{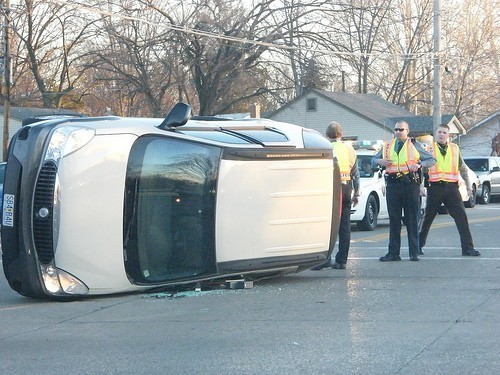} &
    \includegraphics[trim={0 0 0 0}, clip, width=\teaserwidth, height=3cm]{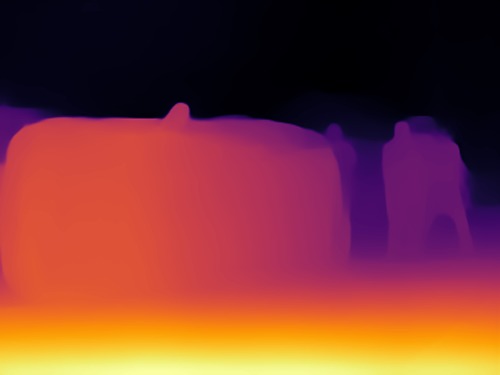} &
     \includegraphics[trim={0 0 0 0}, clip, width=\teaserwidth, height=3cm]{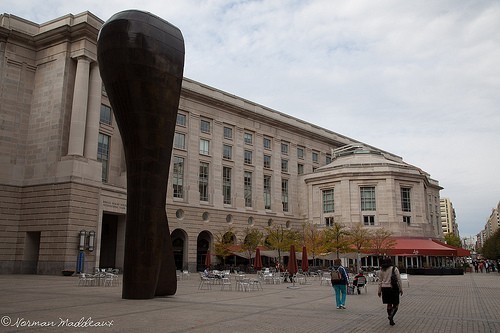} &
    \includegraphics[trim={0 0 0 0}, clip, width=\teaserwidth, height=3cm]{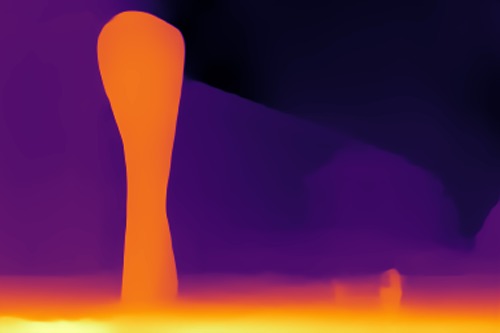} \\
    \end{tabular}
        \caption{Qualitative results on the DIW test set.}
  \label{fig:results_diw}
\end{figure*}

\mypara{Additional qualitative results.}
Figure~\ref{fig:results_diw} shows additional qualitative results on the DIW test set~\cite{Chen2016}. %
We show results on a diverse set of input images depicting various objects and scenes, including humans, mammals, birds, cars, and other man-made and natural objects.
The images feature indoor, street and nature scenes, various lighting conditions, and various camera angles.
Additionally, subject areas vary from close-up to long-range shots.

We show qualitative results on the DAVIS video dataset~\cite{Perazzi2016} in our supplementary video \href{https://youtu.be/D46FzVyL9I8}{\small\texttt{https://youtu.be/D46FzVyL9I8}}.
Note that every frame was processed individually, \ie\ no temporal information was used in any way.
For each clip, the inverse depth maps were jointly scaled and shifted for visualization.
The dataset consists of a diverse set of videos and includes humans, animals, and cars in action.
This dataset was filmed with monocular cameras, hence no ground-truth depth information is available.

\begin{figure*}[htpb]
  \setlength{\linewidth}{\textwidth}
  \setlength{\hsize}{\textwidth}
  \def\mycolspace{1.2mm}
  \def\teaserwidth{0.245\linewidth}

  \centering
 	\begin{tabular}{@{}c@{\hspace{\mycolspace}}c@{\hspace{\mycolspace}}c@{\hspace{\mycolspace}}c@{}}
              \includegraphics[trim={0 0 0 0}, clip, width=\teaserwidth, height=3cm]{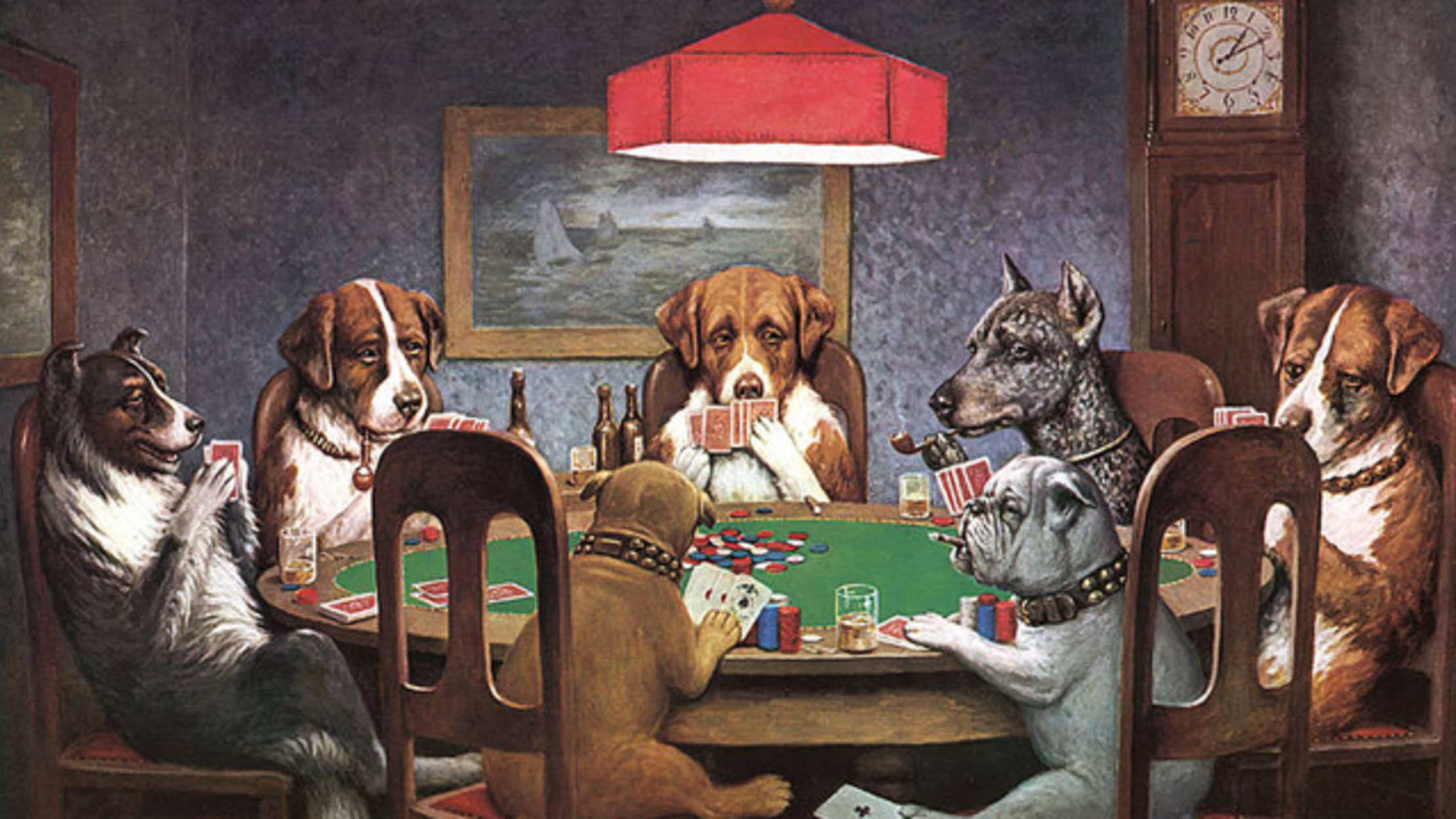} &
              \includegraphics[trim={0 0 0 0}, clip, width=\teaserwidth, height=3cm]{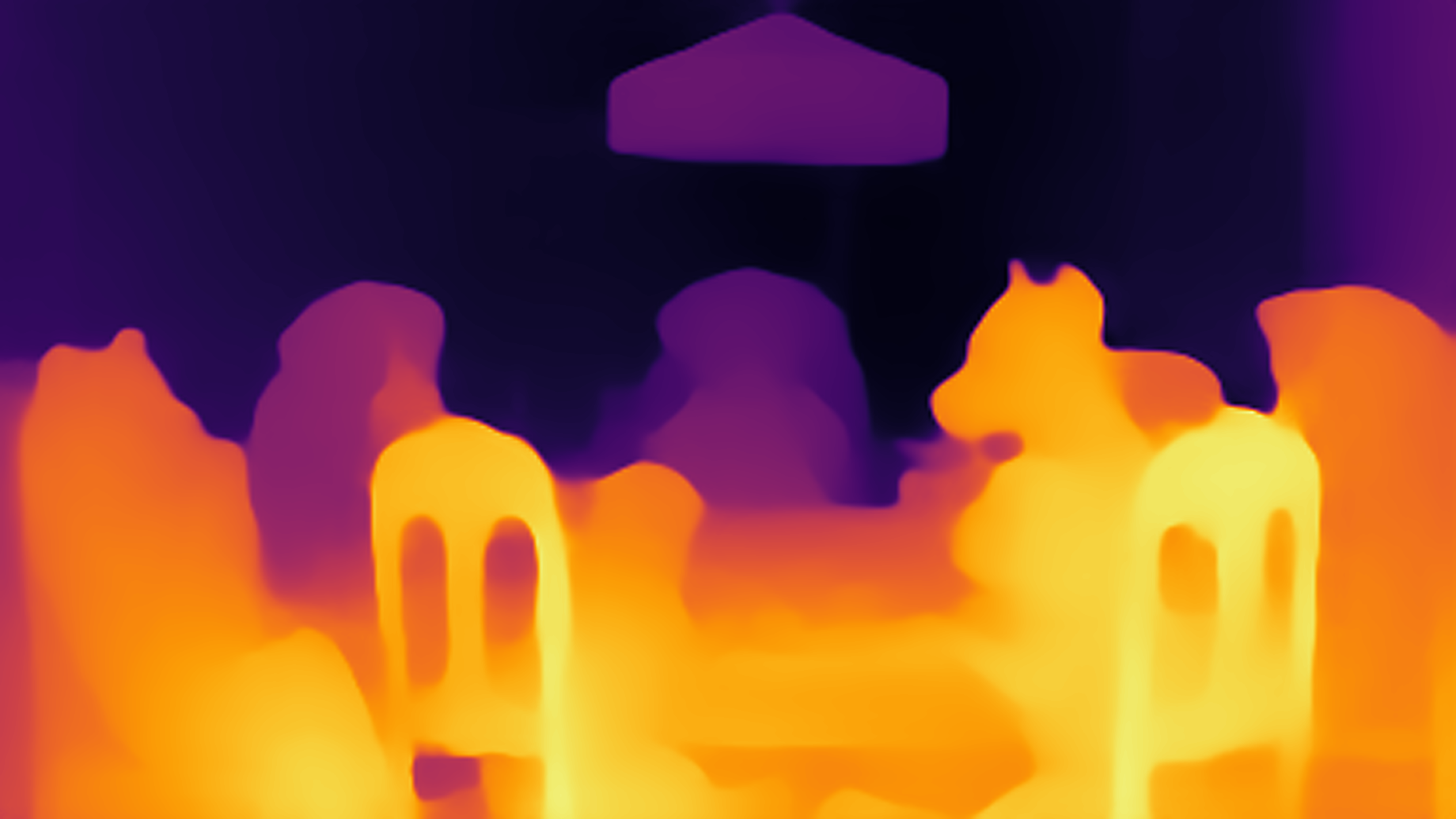} &
              \includegraphics[trim={0 0 0 0}, clip, width=\teaserwidth, height=3cm]{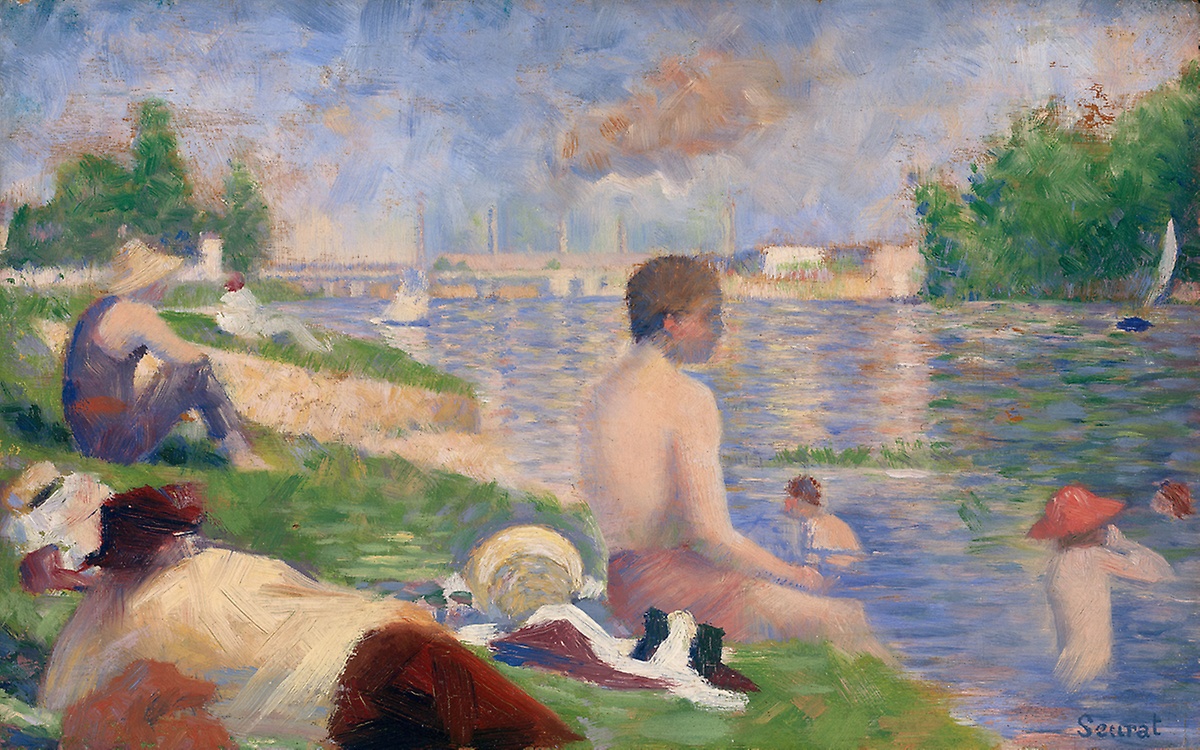} &
              \includegraphics[trim={0 0 0 0}, clip, width=\teaserwidth, height=3cm]{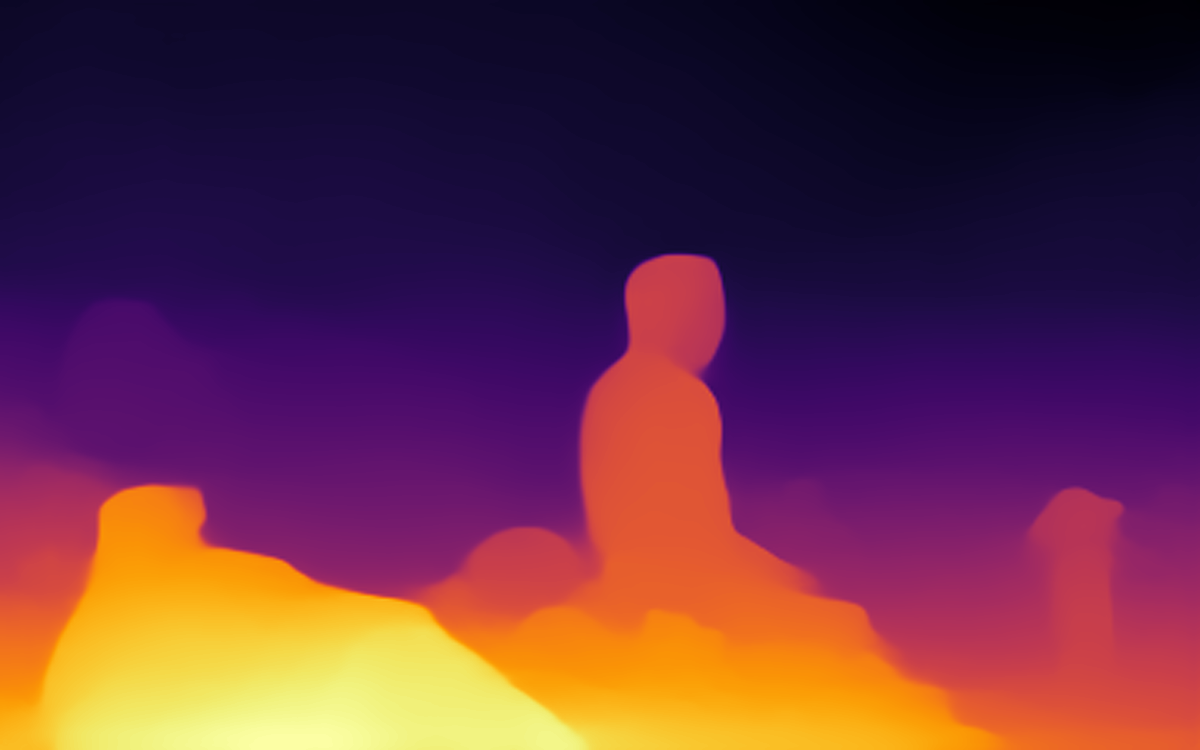} \\
      \includegraphics[trim={0 1.8cm 0 1cm}, clip, width=\teaserwidth]{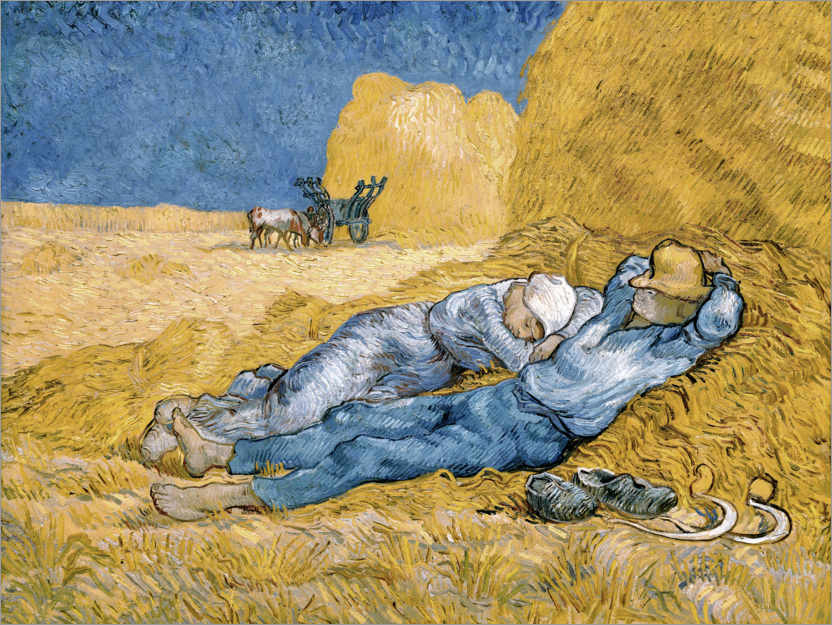} &
              \includegraphics[trim={0 1cm 0 1cm}, clip,  width=\teaserwidth]{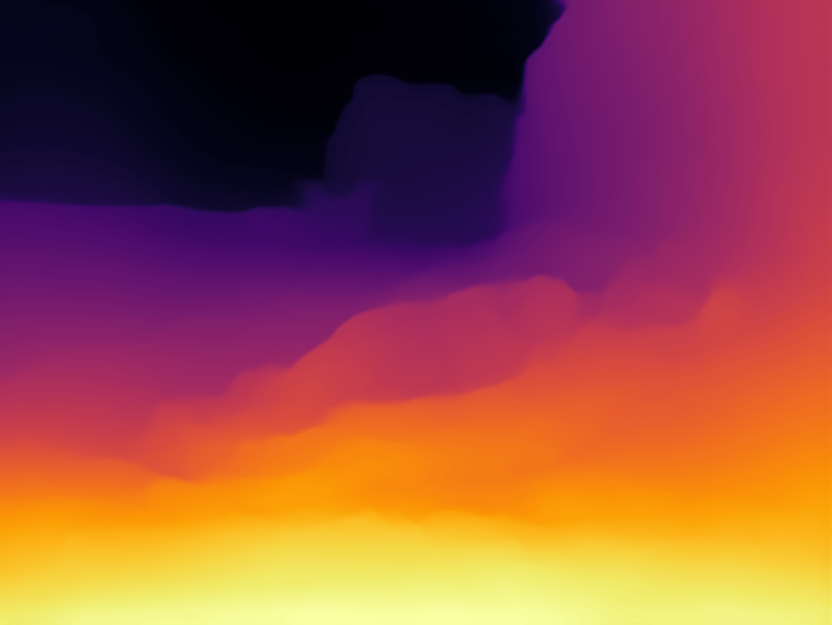} &
              \includegraphics[trim={0 0 0 0}, clip, width=\teaserwidth]{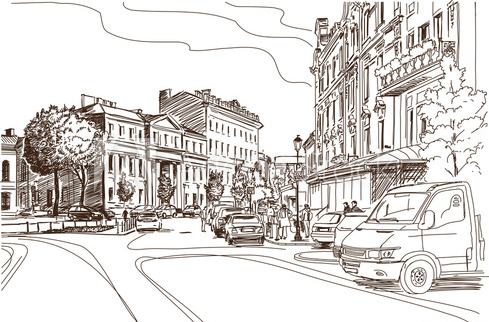} &
              \includegraphics[trim={0 0 0 0}, clip, width=\teaserwidth]{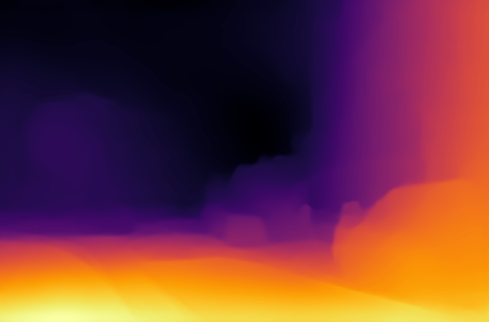}
    \end{tabular}
    \vspace{-0.19cm}
    \caption{Results on paintings and drawings. Top row: \textit{A Friend in Need}, Cassius Marcellus Coolidge, and
      \textit{Bathers at Asni\'{e}res}, Georges Pierre Seurat. Bottom row:  \textit{Mittagsrast}, Vincent van
      Gogh, and  \textit{Vector drawing of central street of old european town, Vilnius, } \href{https://stock.adobe.com/images/vector-drawing-of-central-street-of-old-european-town-vilnius/73034935?asset_id=73034935}{@Misha}}
  \label{fig:art}
\end{figure*}

Hertzmann \cite{Hertzmann2020} recently observed that our publicly available model provides plausible results
even on abstract line drawings. Similarly, we show results on drawings and paintings with different levels of abstraction in
Figure~\ref{fig:art}. We can qualitatively confirm the findings in \cite{Hertzmann2020}: The model shows a
surprising capability to estimate plausible relative depth even on relatively abstract inputs. This seems to be true
as long as some (coarse) depth cues such as shading or vanishing points are present in the artwork.

\begin{figure*}[htpb]

\setlength{\linewidth}{\textwidth}
  \setlength{\hsize}{\textwidth}
  \def\mycolspace{1.2mm}
  \def\teaserwidth{0.245\linewidth}

  \centering
 	\begin{tabular}{@{}c@{\hspace{\mycolspace}}c@{\hspace{\mycolspace}}c@{\hspace{\mycolspace}}c@{}}
              \includegraphics[trim={0 0 0 0}, clip, width=\teaserwidth]{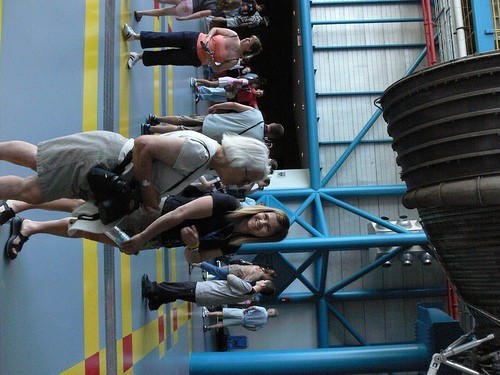} &
              \includegraphics[trim={0 0 0 0}, clip, width=\teaserwidth]{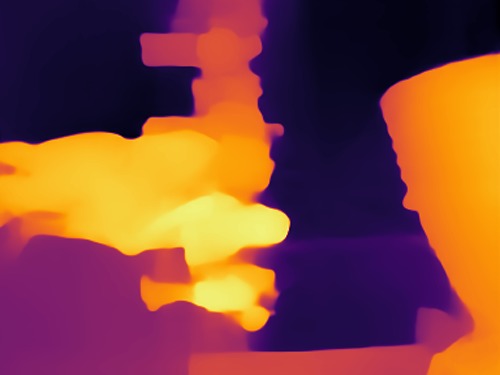} &
              \includegraphics[trim={0 0 0 0}, clip, width=\teaserwidth]{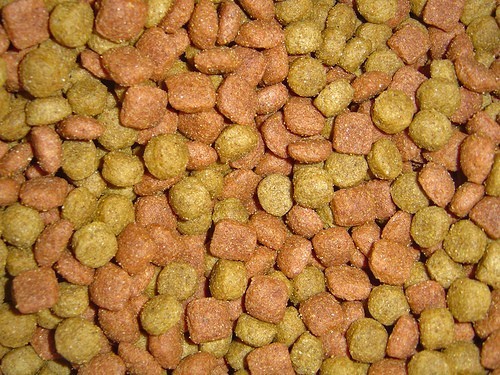} &
              \includegraphics[trim={0 0 0 0}, clip, width=\teaserwidth]{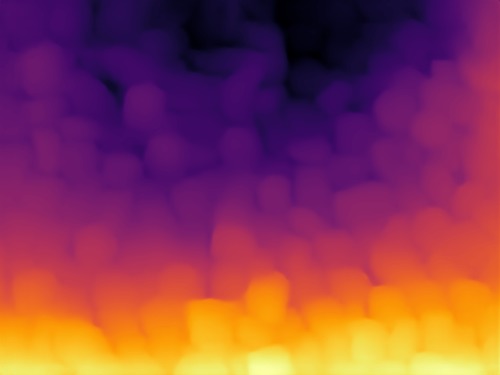} \\
              \includegraphics[trim={0 0 0 1.95cm}, clip, width=\teaserwidth]{figures/results/failure_cases/492_7b2368c17111d22a54dc842638f87e533fc0ef17} &
              \includegraphics[trim={0 0 0 1.95cm}, clip, width=\teaserwidth]{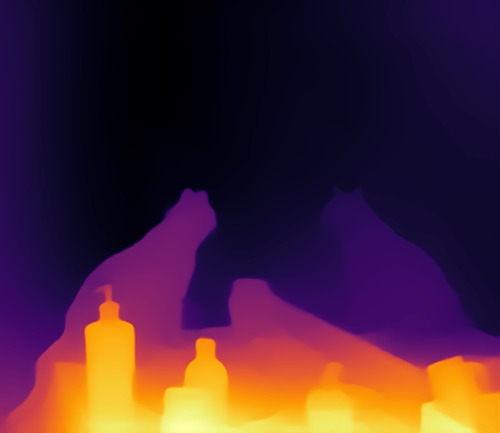} &
              \includegraphics[trim={0 0 0 0}, clip, width=\teaserwidth]{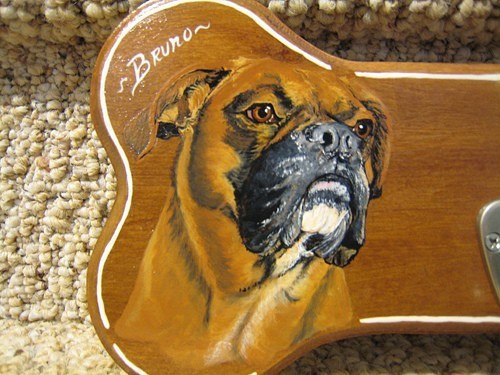} &
              \includegraphics[trim={0 0 0 0}, clip,width=\teaserwidth]{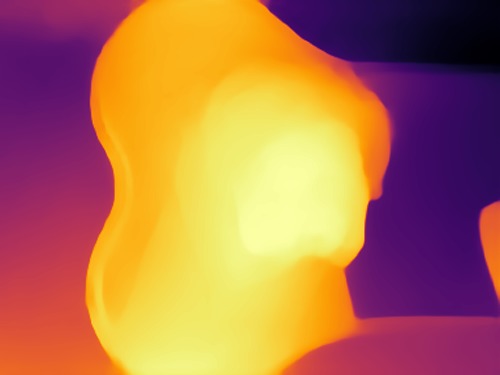} \\
              \includegraphics[trim={0 0 0 2.25cm}, clip, width=\teaserwidth]{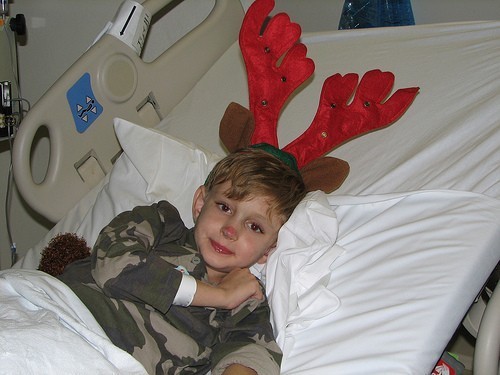} &
              \includegraphics[trim={0 0 0 2.25cm}, clip, width=\teaserwidth]{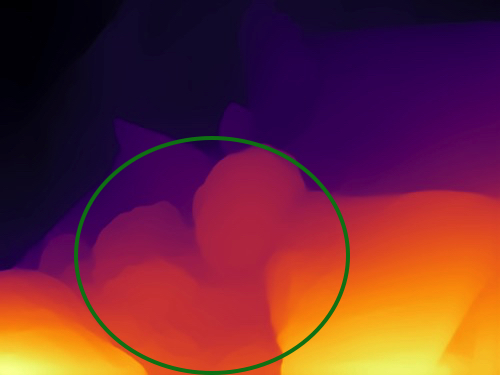} &
              \includegraphics[trim={0 0 0 0}, clip, width=\teaserwidth]{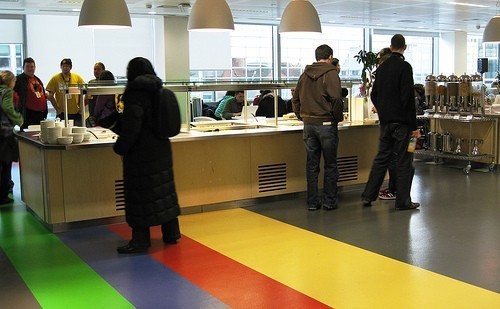} &
              \includegraphics[trim={0 0 0 0}, clip, width=\teaserwidth]{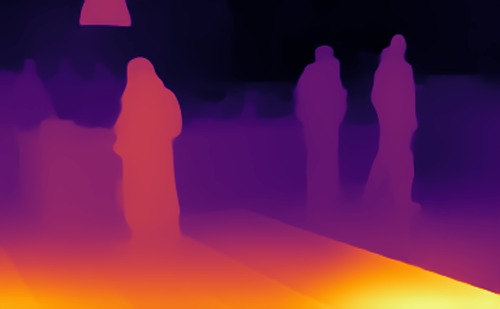} \\
              \includegraphics[trim={0 0 0 0}, clip, width=\teaserwidth]{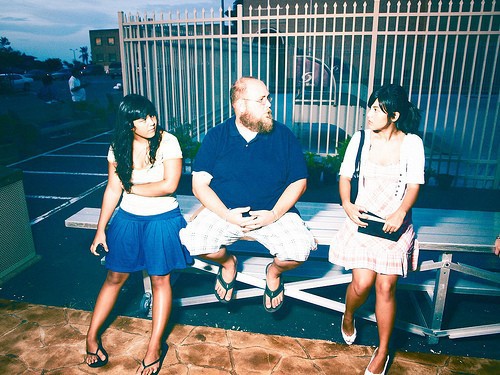} &
              \includegraphics[trim={0 0 0 0}, clip, width=\teaserwidth]{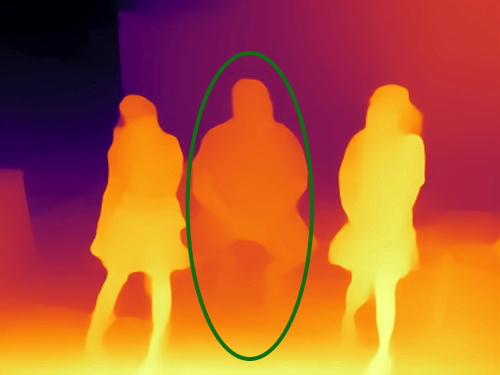} &
              \includegraphics[trim={0 0 0 0}, clip, width=\teaserwidth]{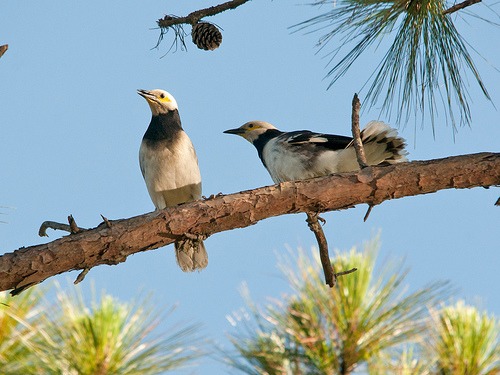} &
              \includegraphics[trim={0 0 0 0}, clip, width=\teaserwidth]{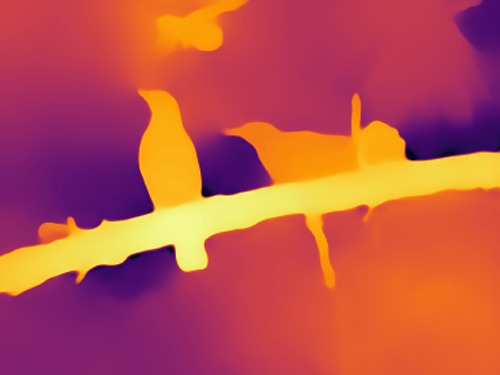} \\
              \includegraphics[trim={0 0 0 0}, clip, width=\teaserwidth]{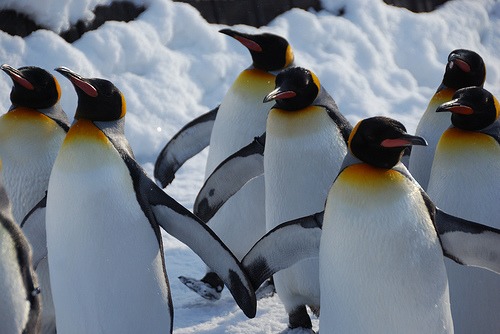} &
              \includegraphics[trim={0 0 0 0}, clip, width=\teaserwidth]{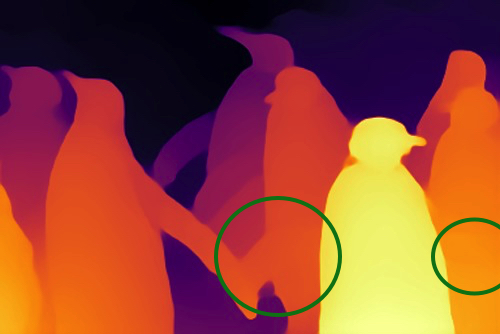} &
              \includegraphics[trim={0 0 0 0}, clip, width=\teaserwidth]{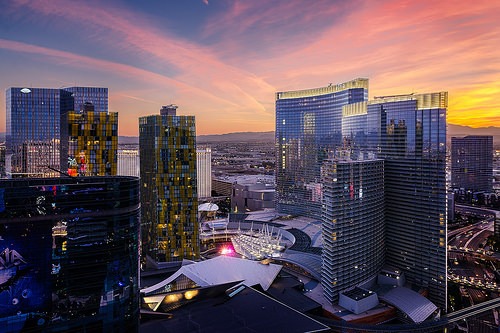} &
              \includegraphics[trim={0 0 0 0}, clip, width=\teaserwidth]{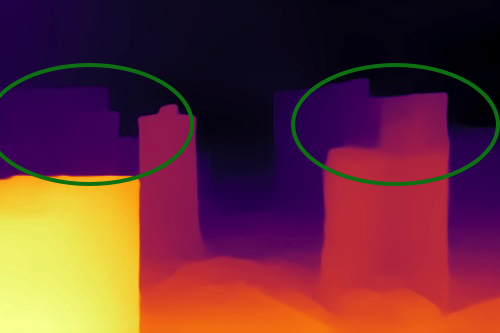}
    \end{tabular}
      \caption{Failure cases. \changed{Subtle failures in relative depth arrangement or missing details are highlighted in green.}}
  \label{fig:fc_big}
\end{figure*}

\mypara{Failure cases.}
We identify common failure cases and biases of our model.
Images have a natural bias where the lower parts of the image are closer to the camera than the higher image regions.
When randomly sampling two points and classifying the lower point as closer to the camera, \cite{Chen2016} achieved an agreement rate of 85.8\% with human annotators.
This bias has also been learned by our network and can be observed in some extreme cases that are shown in the first row of
Figure \ref{fig:fc_big}.
In the example on the left, the model fails to recover the ground plane, likely because the input image was rotated by 90 degrees.
In the right image, pellets at approximately the same distance to the camera are reconstructed closer to the camera in the lower part of the image.
Such cases could be prevented by augmenting training data with rotated images. However, it is not clear if invariance to image rotations is a desired property for this task.

Another interesting failure case is shown in the second row of Figure~\ref{fig:fc_big}. Paintings, photos, and mirrors are often not recognized as
such. %
The network estimates depth based on the content that is depicted on the reflector rather than predicting the depth of the reflector itself.

Additional failure cases are shown in the remaining rows.
Strong edges can lead to hallucinated depth discontinuities. %
Thin structures can be missed and relative depth arrangement between disconnected objects might fail in some situations.
Results tend to get blurred in background regions, which might be explained by the limited resolution of the input images
and imperfect ground truth in the far range.

\begin{table}[!b]
  \caption{Relative performance of state of the art methods with respect to our best model
(top row) -- higher is  better.
Top: models that were not fine-tuned on any of the datasets. Bottom: models that were fine-tuned on a subset of the tested datasets.}
\label{tab:sota}
\centering
\newcolumntype{Q}[0]{S[table-format=3.1,table-number-alignment = center,round-mode=places,round-precision=1,detect-mode]}

	\scalebox{0.86}{
		\ra{1.05} \begin{tabular}{@{}l@{\hspace{0mm}}c@{\hspace{1mm}}|@{\hspace{2mm}}r@{\hspace{2mm}}r@{\hspace{2mm}}r@{\hspace{2mm}}r@{\hspace{2mm}}r@{\hspace{2mm}}r@{\hspace{1mm}}|@{\hspace{1mm}}Q@{}}
	\toprule
	& Training sets &               DIW &              ETH3D &             Sintel &             KITTI &                NYU &                TUM & {Mean [\%]} \\
	\midrule
	Ours & \small \changed{MIX 5} &      \bf{12.46} &         \bf{0.129} &         \bf{0.327} &             23.90 &  \underline{9.55} &         \bf{14.29} & \textemdash\\
    \midrule

                    \changed{Ours -- small} & \small \changed{MIX 5} & \changedr{\underline{-0.2}} & \changedr{\underline{-20.2}}  & \changedr{\underline{-0.9}}   & \changedg{8.7} & \changedr{-64.7} & \changedr{\underline{-19.0}} & \Uline{\changedr{-16.0}}\\
	Xian \cite{Xian2018}  & \small RW &       				 \red{-17.1} & \red{-44.2} &  \red{-29.1} &  \red{-42.6} & \red{-182.7} &  \red{-75.1} &  \red{-65.1} \\
	Li \cite{Li2019}     & \small MC &       				 \red{-112.8} & \red{-41.9} &  \red{-23.9} &  \red{ -100.6} &  \red{-94.5} &  \red{-23.9} &  \red{-66.2} \\
	Wang \cite{Wang2019}  & \small WS &                      \red{-53.2} & \red{-58.9} &  \red{-19.3} &  \red{-33.6} & \red{-209.6} &  \red{-41.2} &  \red{-69.3} \\
	Li \cite{LiSnavely2018}    & \small MD &                \red{-85.8} & \red{-41.1} &  \red{-17.7} &  \red{-51.8} & \red{-188.2} & \red{-106.7} &  \red{-81.9} \\
	Casser \cite{Casser2019}     & \small CS &				 \red{-163.2} & \red{-82.2} &  \red{-29.1} &   \green{11.5} & \red{-314.5} & \red{-160.2} & \red{-122.9} \\
	\midrule
	Fu \cite{Fu2018} &  \small NYU &              			 \red{-131.1} & \red{ -51.2} &  \red{ -32.4} & \red{-157.8} &   \green{\bf{9.0}} &  \red{-72.5} &  \red{-72.6} \\
	Chen \cite{Chen2016}    & \small NYU$\shortrightarrow$DIW & \red{-16.1} & \red{-71.3} &  \red{-34.6} &  \red{-51.9} & \red{-196.6} & \red{-111.1} &  \red{-80.3} \\
	Godard \cite{Godard2019} & \small KITTI &          		 \red{-138.1} & \red{-46.5} &  \red{-24.2} &   \green{\bf{76.9}} & \red{-248.6} & \red{-152.1} &  \red{-88.8} \\
	Casser \cite{Casser2019}    & \small KITTI &     		 \red{-168.8} & \red{-68.2} &  \red{-25.1} &   \green{50.1} & \red{-277.8} & \red{-159.1} &  \red{-108.2} \\
	Fu \cite{Fu2018} & \small KITTI &             			 \red{-143.9} & \red{-67.4} &  \red{-32.1} &   \green{\underline{70.2}} & \red{-325.2} & \red{ -180.8} & \red{-113.2} \\

	\bottomrule
\end{tabular}}
\vspace*{\floatsep}

  \caption{Absolute performance of state of the art methods, sorted by average rank. This table corresponds to Table \ref{tab:sota}.}
  \label{tab:baselines}
  \centering
\scalebox{0.87}{
   \ra{1.05} \begin{tabular}{@{}l@{\hspace{0mm}}c@{\hspace{1mm}}|@{\hspace{1mm}}c@{\hspace{1mm}}c@{\hspace{1mm}}c@{\hspace{1mm}}c@{\hspace{1mm}}c@{\hspace{1mm}}c@{\hspace{1mm}}|@{\hspace{1mm}}c@{}}
	\toprule
	& Training sets &               DIW &              ETH3D &             Sintel &             KITTI &                NYU &                TUM & Rank \\
 & & \changed{WHDR} &  \changed{AbsRel} & \changed{AbsRel} & \changed{$\delta\!\!>\!\!1.25$} & \changed{$\delta\!\!>\!\!1.25$} & \changed{$\delta\!\!>\!\!1.25$} &  \\
	\midrule
    Ours & \small \changed{MIX 5} &      \bf{12.46} &         \bf{0.129} &         \bf{0.327} &             23.90 &  \underline{9.55} &         \bf{14.29} &  \changedt{2.0} \\
    \changedt{Ours -- small} & \small \changed{MIX 5} &  \Uline{\changedt{12.48}} & \Uline{\changedt{0.155}} & \Uline{\changedt{0.330}} & \changedt{21.81}  & \changedt{15.73} & \Uline{\changedt{17.00}} & \changedt{2.7}     \\
    Li  \cite{LiSnavely2018}    & \small MD &             23.15 &  0.181 &  0.385 &             36.29 &              27.52 &              29.54 &  \changedt{5.7} \\
	Li \cite{Li2019}     & \small MC &              26.52 &              0.183 &              0.405 &             47.94 &              18.57 &  17.71 &  \changedt{5.7} \\
	Wang \cite{Wang2019}  & \small WS &         19.09 &              0.205 &              0.390 &             31.92 &              29.57 &              20.18 &  \changedt{6.0} \\
	Xian  \cite{Xian2018} & \small RW &             14.59 &              0.186 &              0.422 &             34.08 &              27.00 &              25.02 &  \changedt{6.1} \\
	 Casser \cite{Casser2019}     & \small CS &             32.80 &              0.235 &              0.422 &             21.15 &              39.58 &              37.18 &  \changedt{9.6} \\
	 \midrule
	 Godard \cite{Godard2019} & \small KITTI &              29.67 &              0.189 &              0.406 &         \bf{5.53} &              33.29 &              36.03 &  \changedt{6.7} \\
	 Fu \cite{Fu2018} & \small NYU &              28.79 &              0.195 &              0.433 &             61.61 &          \bf{8.69} &              24.65 &  \changedt{7.3} \\
	Chen \cite{Chen2016}    &      \small NYU $\shortrightarrow$ DIW &   14.47 &              0.221 &              0.440 &             36.30 &              28.33 &              30.16 &  \changedt{8.5} \\
	Casser \cite{Casser2019}    & \small KITTI &               33.49 &              0.217 &              0.409 &             11.93 &              36.08 &              37.03 &  \changedt{8.7} \\
	Fu \cite{Fu2018} & \small KITTI &             30.39 &              0.216 &              0.432 &  \underline{7.13} &              40.61 &              40.13 &  \changedt{9.2} \\
	\bottomrule

\end{tabular}

}
\end{table}

\section{Conclusion}
\label{sec:conclusion}
The success of deep networks has been driven by massive datasets.
For monocular depth estimation, we believe that existing datasets are still insufficient and likely constitute the limiting factor.
Motivated by the difficulty of capturing diverse depth datasets at scale, we have introduced tools for combining complementary sources of data.
We have proposed a flexible loss function and a principled dataset mixing strategy.
We have further introduced a dataset based on 3D movies that provides dense ground truth for diverse dynamic scenes.

We have evaluated the robustness and generality of models via zero-shot cross-dataset transfer. We find that systematically testing models on datasets that were never seen during training is a better proxy for their performance ``in the wild'' than testing on a held-out portion of even the most diverse datasets that are currently available.

Our work advances the state of the art in generic monocular depth estimation and indicates that the presented ideas substantially improve performance across diverse environments. We hope that this work will contribute to the deployment of monocular depth models that meet the requirements of practical applications.
Our models are freely available at \href{https://github.com/intel-isl/MiDaS}{\small\texttt{https://github.com/intel-isl/MiDaS}}.

\ifCLASSOPTIONcaptionsoff
  \newpage
\fi

\vspace{0.4cm}

\bibliographystyle{IEEEtran}
\bibliography{IEEEabrv,paper}

\begin{IEEEbiography}[{\includegraphics[width=1in,height=1.25in,clip,keepaspectratio]{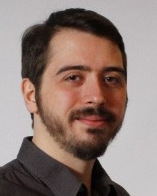}}]{Ren\'{e} Ranftl}
is a Senior Research Scientist at the Intelligent Systems Lab at Intel in Munich, Germany. He received an M.Sc. degree and a Ph.D. degree from Graz University of Technology, Austria, in 2010 and 2015, respectively. His research interests broadly span topics in computer vision, machine learning, and robotics.
\end{IEEEbiography}

\begin{IEEEbiography}[{\includegraphics[width=1in,height=1.25in,clip,keepaspectratio]{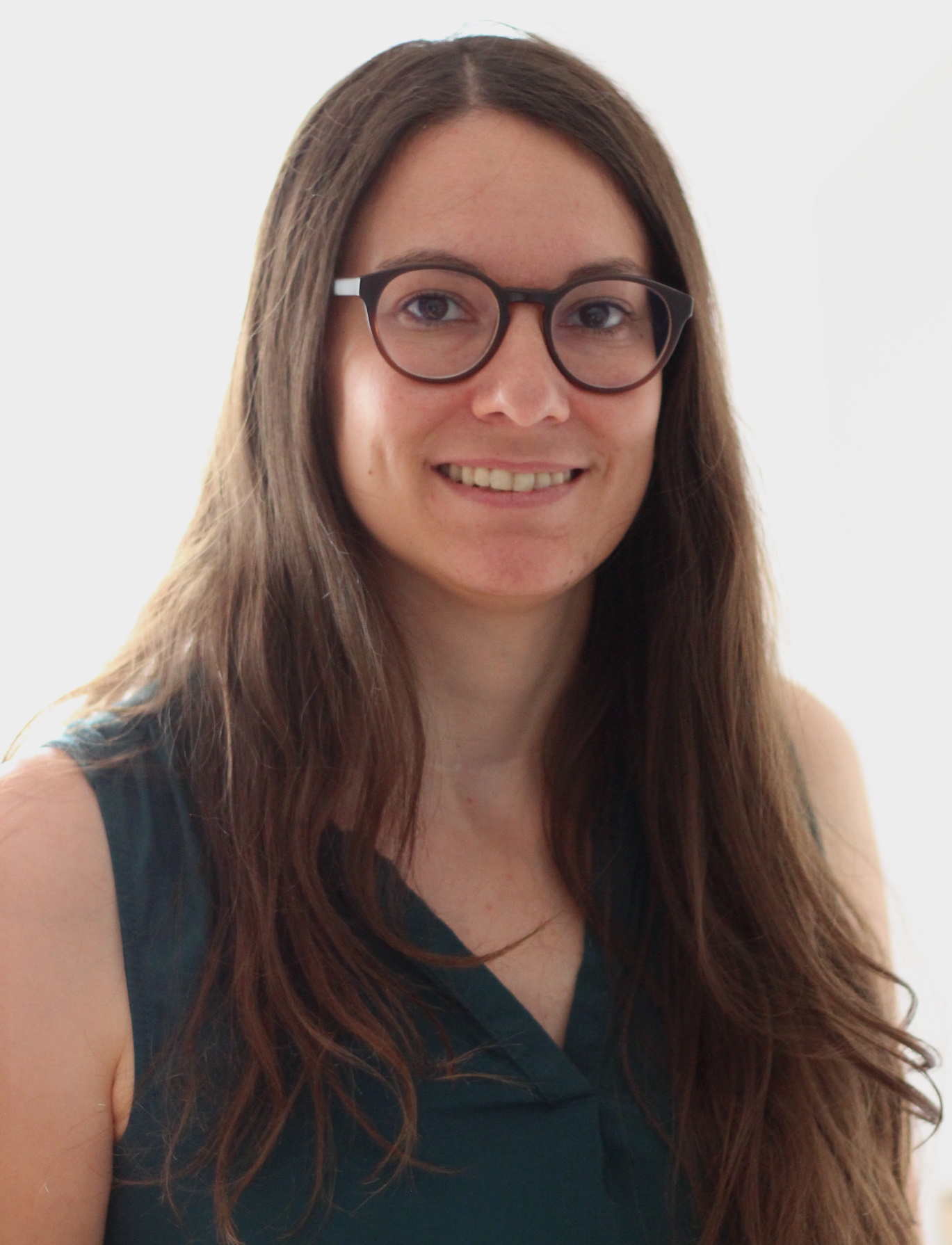}}]{Katrin Lasinger}
received her Master's degree in computer science from TU Wien in 2015. She is currently pursuing her Ph.D. degree in computer vision at the group of Photogrammetry and Remote Sensing at ETH Zurich. Her research is focused on 3D computer vision, including volumetric fluid flow estimation and dense depth estimation from single or multiple views.
\end{IEEEbiography}

\begin{IEEEbiography}[{\includegraphics[width=1in,height=1.25in,clip,keepaspectratio]{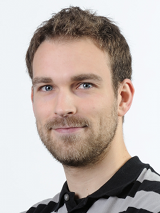}}]{David Hafner}
received a Master's and a Ph.D.\ degree from Saarland University, Germany, in 2012 and 2018, respectively.
Since 2019, he has been a research engineer at the Intelligent Systems Lab at Intel in Munich, Germany.
\end{IEEEbiography}

\begin{IEEEbiography}[{\includegraphics[width=1in,height=1.25in,clip,keepaspectratio]{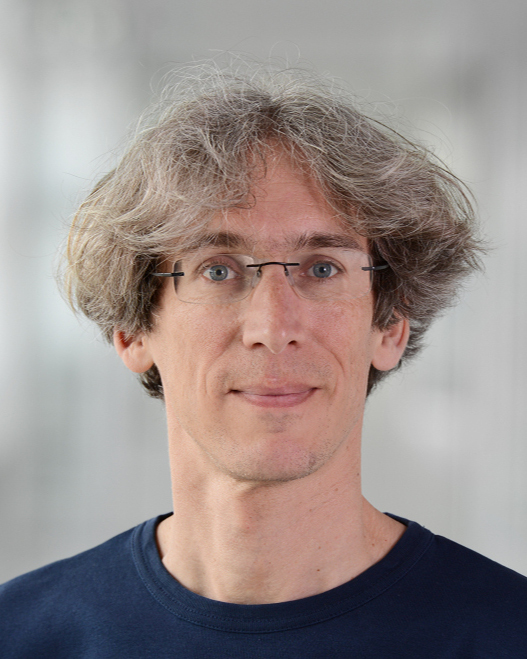}}]{Konrad Schindler}
(M'05–SM'12) received the Diplomingenieur (M.Tech.) degree from Vienna University of Technology, Vienna, Austria, in 1999, and the Ph.D.\ degree from Graz University of Technology, Graz, Austria, in 2003.
He was a Photogrammetric Engineer in the private industry and held researcher positions at Graz University of Technology, Monash University, Melbourne, VIC,
Australia, and ETH Zu\"urich, Z\"urich, Switzerland. He was an Assistant Professor of Image Understanding with TU Darmstadt, Darmstadt, Germany, in 2009. Since 2010, he has been a Tenured Professor of Photogrammetry and Remote Sensing with ETH Z\"urich. His research interests include computer vision, photogrammetry, and remote sensing.
\end{IEEEbiography}

\begin{IEEEbiography}[{\includegraphics[width=1in,height=1.25in,clip,keepaspectratio]{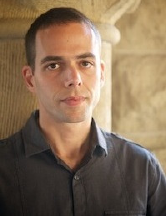}}]{Vladlen Koltun}
is the Chief Scientist for Intelligent Systems at Intel. He directs the Intelligent Systems Lab, which conducts high-impact basic research in computer vision, machine learning, robotics, and related areas. He has mentored more than 50 PhD students, postdocs, research scientists, and PhD student interns, many of whom are now successful research leaders.
\end{IEEEbiography}

\end{document}